%% file: main.tex
\documentclass[10pt,twocolumn,letterpaper]{article}

\usepackage[pagenumbers]{iccv} 

\input{preamble}

\definecolor{iccvblue}{rgb}{0.21,0.49,0.74}
\usepackage[pagebackref,breaklinks,colorlinks,allcolors=iccvblue]{hyperref}

\def\paperID{439} 
\def\confName{ICCV}
\def\confYear{2025}

\title{HyperGCT: A Dynamic Hyper-GNN-Learned Geometric Constraint \\ for 3D Registration}

\author{Xiyu Zhang\textsuperscript{1}, Jiayi Ma\textsuperscript{2}, Jianwei Guo\textsuperscript{3}, Wei Hu\textsuperscript{4}, Zhaoshuai Qi\textsuperscript{1}, Fei Hui\textsuperscript{5}, Jiaqi Yang\textsuperscript{1}\thanks{Corresponding Author. \\ Code link: \url{https://github.com/zhangxy0517/HyperGCT}.}, Yanning Zhang\textsuperscript{1} \\
\textsuperscript{1}Northwestern Polytechnical University \  \textsuperscript{2}Wuhan University \  \textsuperscript{3}Chinese Academy of Sciences \\ \textsuperscript{4}Peking University \ \textsuperscript{5}Chang'an University\\
{\tt\small \{2426988253\}@mail.nwpu.edu.cn; \{ynzhang, jqyang\}@nwpu.edu.cn}
}

\begin{document}
\maketitle
\input{sec/0_abstract}    
\input{sec/1_intro}
\input{sec/2_related}

\input{sec/3_method}
\input{sec/4_exp}
\input{sec/5_con}

{
    \small
    \bibliographystyle{ieeenat_fullname}
    \bibliography{main}
}
\input{sec/X_suppl}
\end{document}

%% file: preamble.tex
\usepackage[dvipsnames]{xcolor}
\usepackage[normalem]{ulem}
\usepackage{gensymb}
\usepackage{amsmath}
\usepackage{multirow}
\usepackage{multicol}
\usepackage{fontawesome}
\usepackage{pifont}
\usepackage{colortbl}
\useunder{\uline}{\ul}{}


%% file: sec/0_abstract.tex
\begin{abstract}
Geometric constraints between feature matches are critical in 3D point cloud registration problems. Existing approaches typically model unordered matches as a consistency graph and sample consistent matches to generate hypotheses. However, explicit graph construction introduces noise, posing great challenges for handcrafted geometric constraints to render consistency. To overcome this, we propose HyperGCT, a flexible dynamic \textbf{Hyper}-\textbf{G}NN-learned geometric \textbf{C}onstrain\textbf{T} that leverages high-order consistency among 3D correspondences. To our knowledge, HyperGCT is the first method that mines robust geometric constraints from dynamic hypergraphs for 3D registration. By dynamically optimizing the hypergraph through vertex and edge feature aggregation, HyperGCT effectively captures the correlations among correspondences, leading to accurate hypothesis generation. Extensive experiments on 3DMatch, 3DLoMatch, KITTI-LC, and ETH show that HyperGCT achieves state-of-the-art performance. Furthermore, HyperGCT is robust to graph noise, demonstrating a significant advantage in terms of generalization. 
\end{abstract}

%% file: sec/1_intro.tex
\section{Introduction}
\begin{figure}[t]
  \centering
   \includegraphics[width=\linewidth]{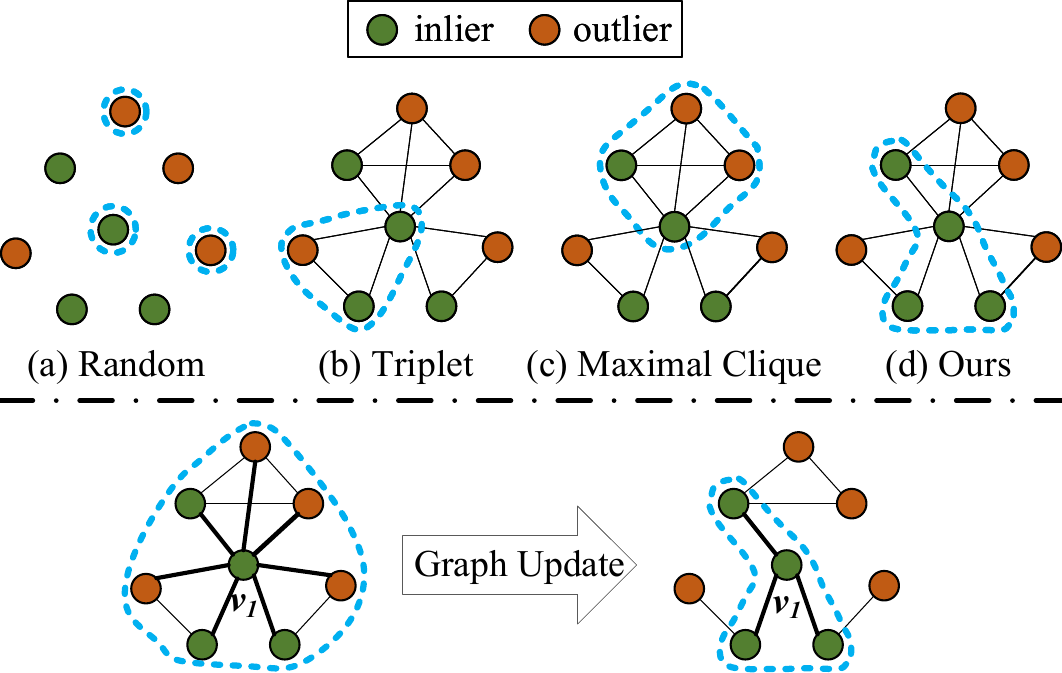}
   \caption{Key aspects of HyperGCT. It mines high-order consistency among inliers, offering greater flexibility than random sampling, triplet, or clique constraints. The implicit graph construction continuously refines the neighbor connections around the central node, effectively learning the high-order correlations.}
   \label{fig:fig1}
\end{figure}
Point cloud registration (PCR) is essential in numerous 3D computer vision tasks including reconstruction~\cite{choi2015robust}, simultaneous localization and mapping (SLAM)~\cite{durrant2006simultaneous}, and robotics~\cite{pomerleau2015review}. It aims to align 3D fragments from different views. The standard approach involves establishing correspondences using local features, sampling consistent matches that meet geometric constraints, and estimating six-degree-of-freedom (6-DoF) pose transformations. However, PCR becomes challenging when faced with noise, such as low-quality point clouds or incorrect matches.

Correspondence-based methods are the mainstream solution to PCR and can be categorized into deep-learned and traditional approaches. Deep-learned methods~\cite{choy2020deep,pais20203dregnet,bai2021pointdsc,lee2021deep,wang2023pg,jiang2023robust,yao2023hunter,wang20243dpcp} classify inliers and outliers by learning the consistency between correspondences and generating inlier probabilities for outlier removal. Most traditional methods~\cite{yang2021sac, yang2022correspondence, sun2022trivoc, wu2022robust, yang2023mutual, cheng2023sampling, yang2020teaser, li2024effective, yang2024mac, qiao2024g3reg} follow the hypothesis generation-verification paradigm, leveraging geometric constraints to select consistent matches. Thus, designing a good geometric constraint is critical for both methods. Deep-learned methods integrate pairwise compatibility into networks, for example, PointDSC~\cite{bai2021pointdsc} using spatial consistency and 3DPCP-Net~\cite{wang20243dpcp} incorporating deep geometric coherence. Traditional methods employ random sampling, triplet, or clique constraints. RANSAC~\cite{fischler1981random} and its variants~\cite{chum2005matching, ni2009groupsac, barath2018graph, barath2020magsac++} use random sampling to remove outliers, but their performance degrades with a high outlier ratio. Recently, researchers~\cite{yang2021sac, yang2022correspondence, sun2022trivoc, wu2022robust, yang2023mutual,yang2020teaser, li2024effective, yang2024mac, qiao2024g3reg} have applied graph theory to identify geometrically consistent matches. These approaches typically check the geometric consistency with empirical thresholds between correspondences before explicitly constructing a graph to represent the relationships. SAC-COT~\cite{yang2021sac} leverages triplet sampling to accelerate the process and enhance efficiency and accuracy. Assuming that inliers form a maximum clique, TEASER~\cite{yang2020teaser} uses a parallel maximum clique finder algorithm, PMC~\cite{rossi2015parallel}, to identify this clique and filter out outliers. MAC~\cite{yang2024mac} relaxes the maximum clique constraints by searching for maximal cliques, fully considering each local consensus. However, existing constraints have several key issues: \textbf{1)} pairwise constraint is ambiguous and struggles to associate multiple correspondences; \textbf{2)} static compatibility thresholds limit the flexibility to handle various input conditions; \textbf{3)} explicit graph construction is sensitive to graph noise introduced by matching and measurement errors.

In this paper, to address the above problems in PCR, we propose HyperGCT. This dynamic Hyper-GNN-learned geometric constraint leverages high-order consistency to capture the relationships among correspondences in a latent hypergraph feature space. The motivation is that measurement and matching errors can make geometric constraints in explicit space highly vulnerable to graph noise. By representing correspondences within a hypergraph, HyperGCT naturally captures complex, higher-order relationships that are more robust to noise than standard graph-based approaches. As shown in Fig.~\ref{fig:fig1}, HyperGCT differs from others in several key aspects. First, unlike other methods, HyperGCT uses a hypergraph to model multiple correspondences, offering more flexibility than random sampling, triplet, or clique constraints. With implicit graph construction and continuous topology updates via Hyper-GNN~\cite{feng2019hypergraph}, our approach is parameter-insensitive and robust to graph noise. Second, unlike methods focus on inlier/outlier classification, HyperGCT emphasizes learning high-order correlations among correspondences, enabling it to handle matches across different modalities (RGB-D and LiDAR), and providing stronger generalization than existing approaches. Overall, our contributions are as follows:
\begin{itemize}
    \item We introduce HyperGCT, a novel geometric constraint for 3D correspondences that offers greater flexibility than existing constraints, \eg, random sampling, triplet, and maximal cliques.
    \item We further propose a HyperGCT-based registration network. Our method demonstrates a significant advantage in registration performance and generalization compared to other approaches. 
\end{itemize}

%% file: sec/2_related.tex
\begin{figure*}[ht]
\centering
    \includegraphics[width=\linewidth]{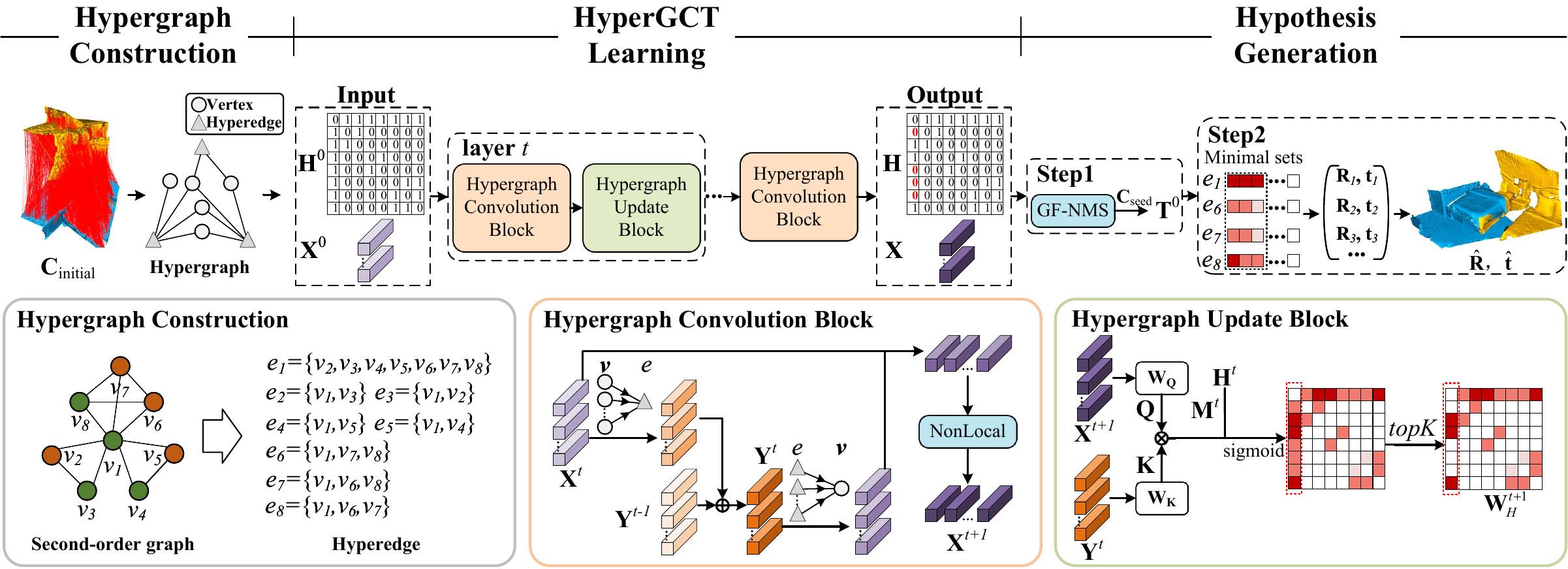}
    \caption{{\textbf{The overall pipeline}}. \textbf{1.} An initial hypergraph ${\bf H}^0$ encodes high-order consistency within ${\bf C}_{\text{initial}}$ \textbf{2.} Hypergraph is continuously updated through HyperGCT learning.  \textbf{3.} Hypotheses generated from high-order subsets are evaluated to get the final transformation.}
    \label{fig:pipline}
\end{figure*}
\section{Related Works}
\subsection{Point Cloud Registration}
\noindent\textbf{Traditional Methods.} The RANSAC algorithm~\cite{fischler1981random} is widely employed in PCR following the hypothesis generation-verification process.  Over the years, many variants have been proposed to improve sampling~\cite{torr2002napsac,chum2005matching, ni2009groupsac,barath2020magsac++} and verification~\cite{torr2000mlesac, chum2003locally, chum2008optimal,barath2018graph, yang2021toward}. However, due to random sampling and the lack of inlier consistency constraints, these methods struggle with efficiency and accuracy when the outlier ratio is high. To address this, researchers model the unordered correspondences as a compatibility graph and use geometric constraints to guide hypothesis generation. Some approaches~\cite{yang2021sac,yang2022correspondence,sun2022trivoc,wu2022robust,yang2023mutual} utilize triplet constraints to reduce the ambiguity of pairwise consistency. Assuming that inliers are mutually compatible with each other, some approaches~\cite{yang2020teaser,li2024effective,yang2024mac,qiao2024g3reg} utilize clique constraints. 

Traditional methods face two key issues: they require manual parameter tuning, limiting flexibility when input conditions change, and their reliance on explicit graph construction makes them sensitive to graph noise from matching and measurement errors. In contrast, we use a hypergraph to model relationships among multiple correspondences, offering greater flexibility than other constraints. With implicit graph construction and continuous topology updates via Hyper-GNN, our method is parameter-insensitive and robust to graph noise.

\noindent\textbf{Deep-learned Methods.} These methods~\cite{choy2020deep,pais20203dregnet,bai2021pointdsc,lee2021deep,wang2023pg,jiang2023robust,yao2023hunter,wang20243dpcp} design inlier/outlier classifiers by learning consistency between correspondences. DGR~\cite{choy2020deep} and 3DRegNet~\cite{pais20203dregnet} employ sparse convolution and point-by-point MLP to train end-to-end neural networks. DHVR~\cite{lee2021deep} leverages deep Hough voting to establish consensus in the Hough space. PG-Net~\cite{wang2023pg} proposes a grouped dense fusion attention feature embedding module to enhance inlier representations. VBReg~\cite{jiang2023robust} models long-range dependencies with a variational non-local network. However, they often neglect geometric consistency constraints. PointDSC~\cite{bai2021pointdsc} incorporates spatial consistency into a NonLocal module. 3DPCP-Net~\cite{wang20243dpcp} introduces a DGC block to explore deep feature similarity, pairwise distances, and angles. Despite these advancements, they remain limited to refining higher-order relationships beyond pairwise vertex interactions. Though Hunter~\cite{yao2023hunter} addresses this limitation by introducing an HCR module to learn the high-order consistency of inliers, its hypergraph structure is static and lacks flexibility. 

Deep-learned methods generally outperform traditional approaches but struggle with high outlier ratios due to weak inlier consistency. Furthermore, reliance on data-specific features hinders generalization in varying pattern matching of characteristics in different datasets. In contrast, we operate within a generalized hypergraph framework to optimize the relationships between vertices and hyperedges, enabling enhanced, precise information propagation and robust modeling of higher-order relationships. Our method effectively handles diverse matching patterns while improving generalization across various scenarios.

\subsection{Hypergraph Learning for Point Clouds}
A hypergraph extends graph theory by modeling complex relationships. Unlike regular graphs, where an edge connects two nodes, hypergraphs use hyperedges to connect two or more nodes, capturing higher-order relationships. This is particularly useful for high-dimensional data, such as point clouds, where higher-order correlations exist. Hypergraphs improve the model's expressiveness and accuracy, making them valuable for point cloud processing. In recent years, hypergraph learning has been applied to tasks like  sampling~\cite{zhang2020hypergraph,deng2022efficient}, denoising~\cite{zhang2020hypergraph}, classification~\cite{zhang2018inductive}, recognition~\cite{bai2021multi}, and registration~\cite{yao2023hunter}.

%% file: sec/3_method.tex
\section{Method}
\subsection{Problem Formulation}
Given the source point cloud ${\bf P}^s=\{{\bf p}^s\}$ and target point cloud ${\bf P}^t=\{{\bf p}^t\}$ to be aligned, we first extract local features for them using geometric~\cite{rusu2009fast} or learned~\cite{choy2019fully} descriptors. An initial correspondence set ${\bf C}_{\text{initial}}=\{{\bf c}\}$ is then established by nearest neighbor search in feature space, where ${\bf c}=({\bf p}^s,{\bf p}^t)$. The final goal is to estimate an accurate pose transformation from ${\bf C}_{\text{initial}}$. To achieve this, we propose HyperGCT, designed to capture and represent high-order consistency among inliers for each correspondence. As illustrated in Fig.~\ref{fig:pipline}, our method comprises three stages: hypergraph construction, HyperGCT learning, and hypothesis generation. 1) \emph{Hypergraph construction}: we build an initial hypergraph that encodes high-order consistency within ${\bf C}_{\text{initial}}$, allowing HyperGCT to optimize the structure. 2) \emph{HyperGCT learning}: to avoid generalization issues from simply learning to distinguish inliers from outliers, we leverage inliers' high-order consistency to dynamically update the hypergraph structure and refine correspondence representation, which helps to improve the reliability of hypothesis generation. 3) \emph{Hypothesis generation}: this final stage samples putative seed correspondences from the updated hypergraph, generating and evaluating multiple hypotheses. The best hypothesis is selected as the final transformation estimate.

\subsection{Hypergraph Construction}\label{sec:graph_con}
A hypergraph is defined as $\mathcal{H}=(\pmb{\mathcal{V}},\pmb{\mathcal{E}},{\bf H})$, where $\pmb{\mathcal{V}}$ is the set of vertices and $\pmb{\mathcal{E}}$ is the set of hyperedges. The incidence matrix ${\bf H} \in {\{ 0,1\} ^{\left| {\pmb{\mathcal{V}}} \right| \times \left| {\pmb{\mathcal{E}}} \right|}}$ represents hypergraph's structure, with ${\bf H}(v_i,e_j)$ indicating whether hyperedge $e_j$ contains vertex $v_i$. The degree of hyperedge $e_j$ and vertex $v_i$ is defined as ${\bf{D}}({e_j}) = \sum\limits_{i = 1}^{\left| \mathcal{V} \right|} {{\bf{H}}({v_i},{e_j})}$ and ${\bf{D}}({v_i}) = \sum\limits_{j = 1}^{\left| \mathcal{E} \right|} {{\bf{H}}({v_i},{e_j})}$. Unlike an edge in a simple graph that connects only two vertices, a hyperedge connects multiple vertices, capturing high-order relationships. We
model ${\bf C}_{\text{initial}}$ as an initial hypergraph $\mathcal{H}_0$, with each vertex $v$ representing a correspondence $\bf c$. To construct hyperedges, we use the second-order measure~\cite{chen2023sc, yang2024mac}, which has a stronger constraint capability than the first-order measure (\ie, pairwise measure). As illustrated in Fig.~\ref{fig:pipline}, we first model ${\bf C}_{\text{initial}}$ as a second-order graph (SOG) based on the pairwise rigid distance compatibility measure. For a correspondence pair (${\bf c}_i,{\bf c}_j$), the rigid distance is:
\begin{equation}
    {d_{ij}} = \left| {\left\| {{{\bf p}}_i^s - {{\bf p}}_j^s} \right\| - \left\| {{{\bf p}}_i^t - {{\bf p}}_j^t} \right\|} \right|.
\end{equation}
The compatibility score is:
\begin{equation}
    \gamma_{ij} =\left[ 1 - \frac{{{d_{ij}}^2}}{{{\sigma _\text{d}}^2}}\right]_+,
\end{equation}
where ${\sigma _\text{d}}$ controls the sensitivity to distance difference. $\left[\cdot\right]_+$ is the operation of $max(\cdot,0)$. Usually, 
if $\gamma_{ij}$ exceeds threshold $\theta_\text{cmp}$, the two correspondences are considered compatible, and $\gamma_{ij}$ is stored in matrix ${\bf W}_{\gamma}$. Different from methods that use static thresholds~\cite{yang2021sac,yang2024mac, bai2021pointdsc, chen2023sc}, we compute $\theta_\text{cmp}$ in a dynamic way to handle ${\bf C}_{\text{initial}}$ with various inlier ratios:
\begin{equation}
    {\theta _\text{cmp}} = \frac{1}{{K_1\left| {{{\bf C}_{\text{initial}}}} \right|}}\sum\limits_{i=0}^{\left|{\bf C}_{\text{initial}}\right|} {\sum\limits_{j=0}^{K_1} {topK_1(\gamma_{ij})} },
\end{equation}
where $K_1=0.1\left| {\bf C}_{\text{initial}}\right|$.
The initial weight matrix ${\bf W}_H^0$ is:
\begin{equation}
    {{{\bf W}_H^0}} = {{{\bf W}}_{\gamma}} \odot ({{{\bf W}}_{\gamma}} \times {{{\bf W}}_{\gamma}}),
\end{equation}
where $\odot$ denotes the element-wise product. Next, we define the vertices directly connected to each non-isolated vertex $v_i$, in the SOG as belonging to the hyperedge $e_i$.  For vertices $v_i$ and $v_j$, ${\bf H}(v_i,e_j)$ is set to 1 if ${{\bf W}}_H(i,j)>0$, and the initial hypergraph ${\bf H}^0$ is constructed. Thus, $e_i$ contains the high-order consistency information surrounding $v_i$, {serving as the geometric constraint optimized in the HyperGCT learning step}. 

\subsection{HyperGCT Learning} {This step  optimizes each hyperedge $e_i$ based on initial hypergraph ${\bf H}^0$, ensuring it includes only the inliers. High-order consistency information is then leveraged to represent $v_i$ accurately.}
Similar to traditional GNNs~\cite{lin2021learning}, HyperGCT employs a stacked structure of multiple convolution layers to learn feature representations progressively. Each convolution layer consists of a hypergraph convolution block, updating vertex and hyperedge features and enabling extensive information exchange. Between adjacent hypergraph convolution blocks, a hypergraph update block is introduced. This block refines the graph structure by assessing the similarity between the features of vertices and hyperedges, ensuring the structural relationships are continuously optimized. Combining the two kinds of blocks allows HyperGCT to capture high-order consistency among inliers effectively. The coordinates of correspondences are taken as the network's input, making our method applicable to different descriptors. 

\noindent\textbf{Hypergraph Convolution Block.} 
The information passing in the $t$-th hypergraph convolution block is divided into two stages: the first stage aggregates the information of vertices within each hyperedge to update the hyperedge features from ${\bf Y}_{t-1}$ to ${\bf Y}_t$, while the second stage aggregates the hyperedge information to update the vertex features from ${\bf X}_t$ to ${\bf X}_{t+1}$.
The first stage is formulated as follows:
\begin{equation}
    Stage1\left\{ \begin{array}{l}
{{{\bf{\hat Y}}}^t}=({\bf D}_e^t)^{-1}{({{\bf{H}}^t})^\top}{{\bf{X}}^t}\\
{{\bf{Y}}^t} = \text{normalize}(MLP_1(concate({{\bf{Y}}^{t - 1}},{{{\bf{\hat Y}}}^t})))
\end{array} \right.,
\end{equation}
where ${\bf D}_e^t$ is the diagonal matrix of hyperedge degrees ${\bf D}(e)^t$. ${\bf Y}^t$ is obtained by merging the aggregated vertex information ${{\bf{\hat Y}}}^t$ with ${\bf Y}^{t-1}$ obtained from the previous block. The motivation is to enhance the feature representation by combining the local contextual information and historical feature representations, thereby more accurately capturing the high-order consistency. 
The second stage is formulated as follows:
\begin{equation}
    Stage2\left\{ \begin{array}{l}
{{{\bf{\hat X}}}^{t + 1}}{\rm{ = ({\bf D}}}{}_v^t{{\rm{)}}^{ - 1}}{{\bf{H}}^t}{\bf{W}}_{\rm{e}}^t{{\bf{Y}}^t}\\
{{{\bf{X}}}^{t + 1}} = \rho ({{\bf{X}}^t} + MLP_2({{{\bf{\hat X}}}^{t + 1}}))\\
{{\bf{X}}^{t + 1}} = \text{normalize} (NonLocal({{{\bf{X}}}^{t + 1}},{{\bf{W}}_H^0}))
\end{array} \right.,
\end{equation}
where $\rho$ represents the ReLU activation function, ${\bf D}_v^t$ is the diagonal matrix of vertex degrees ${\bf D}(v)^t$, and ${\bf W}_e^t$ is the diagonal matrix of hyperedge weights ${\bf W}(e)^t$, definded as:
\begin{equation}
    {{{\bf W}(}}{{\rm{e}}_j}{{\rm{)}}^t}{\rm{ = }}\sum\limits_{i = 1}^{\left| \mathcal{V} \right|} {{\bf{W}}_H^t(i,j)}.
\end{equation}

When updating the vertex feature ${\bf X}^{t+1}$ using the aggregated hyperedge information ${{\bf{\hat X}}}^{t + 1}$, we introduce a self-looping message passing path to prevent the vanishing of the vertex's own feature information~\cite{huang2021unignn}. Subsequently, the vertex feature ${\bf X}^{t+1}$ is passed through a NonLocal~\cite{bai2021pointdsc} block, where feature enhancement is performed using the matrix ${\bf W}_H^0$, which contains the global second-order compatibility information.

\noindent\textbf{Hypergraph Update Block.} The block first generates a mask matrix ${\bf M}^t$ based on ${\bf H}^t$, which is used to ignore vertices and hyperedges with no connections:
\begin{equation}
    {{\bf{M}}^{t}} = (1 - {{\bf{H}}^t})\cdot(-\infty).
\end{equation}
Next, ${{\bf{H}}^{t + 1}}$ and ${\bf{W}}_H^{t + 1}$ are obtained through the graph pooling operation based on the similarity between vertex feature ${\bf X}^{t+1}$ and hyperedge feature ${\bf Y}^t$: 
\begin{equation}
    {{\bf{H}}^{t + 1}},{\bf{W}}_H^{t + 1} = topK_2(sigmoid(\frac{{{({{\bf Q}^t})^\top}{\bf K}^t}}{{\sqrt {{{N}_\text{channel}}} }} + {{\bf{M}}^{t}})),
\end{equation}
where $K_2=0.1(5-t)\left|{\bf C}_{\text{initial}}\right|$, $t \in \left[0,4\right]$ and ${N}_\text{channel}$ is the number of feature channels. ${\bf Q}^t$ and ${\bf K}^t$ are the projections of ${\bf X}^{t+1}$ and ${\bf Y}^t$. In this step, each vertex selects ${ K}_2$ most similar hyperedges in feature space, ensuring that the selected hyperedges contain the vertex. 

As $K_2$ decreases with increasing $t$, deeper layers allow each vertex to select fewer hyperedges. This encourages more comprehensive information exchange at shallower layers, while deeper layers focus on interactions among inliers. Because the similarity is computed between each vertex-hyperedge pair, the sigmoid activation function is applied here. The learned feature ${\bf X}_i$ for each correspondence and optimized incidence matrix ${\bf H}$ are used for 
 predicting the inlier confidence  $\hat{\bf s}$ and hypothesis generation. 
\subsection{Hypothesis Generation}
The hypotheses are generated through guided sampling, which contains two steps. In the first step, initial hypotheses are derived from putative seeds. Leveraging both inlier confidence $\hat{\bf s}$ and the hypergraph structure ${\bf H}$, we propose a Graph-Filtering Non-Maximum Suppression (GF-NMS) method to retrieve more reliable correspondences as seeds ${\bf C}_{\text{seed}}$ compared to standard NMS~\cite{lowe2004distinctive} (refer to Supp.~\ref{sec:GF_NMS}):
\begin{equation}
    {\bf C}_{\text{seed}}=\text{GF-NMS}({\bf H}, \hat{\bf s},N_\text{s}),
\end{equation}
where $N_\text{s}$ is the number of seeds. A feature-space KNN search is then conducted to sample subsets for each correspondence in ${\bf C}_{\text{seed}}$. Subsequently, the SVD algorithm is applied to each subset to generate candidate hypotheses, with $N_{\text{initial}}$ of the most promising ones retained as initial hypotheses ${\bf T}^0$.

Since ${\bf T}^0$ may not obtain local optima, the second step is introduced to explore the solution space further, guided by high-order consistency, to find the optimal hypothesis. Each ${\bf c}_i$ in ${\bf C}_{\text{seed}}$ corresponds to a hyperedge ${e}_i$ in $\bf H$, which includes multiple correspondences exhibiting high-order consistency with ${\bf c}_i$. We compute the residuals for correspondences within ${ e}_i$ using the initial hypothesis ${\bf T}^0_i$ and sort them in ascending order based on the residual value:
    $r_{ij}=\left\| {{{\bf{R}}_i}{\bf{p}}{}_{ij}^s + {{\bf{t}}_i} - {\bf{p}}{}_{ij}^t} \right\|$, where ${\bf c}_{ij}=({\bf{p}}_{ij}^s,{\bf{p}}_{ij}^t)$ represents a correspondence in ${ e}_i$.
For each hyperedge $e$, we iteratively sample minimal sets from front to back to generate hypotheses. Early-stage samples produce hypotheses similar to the initial ones, while later-stage samples generate hypotheses that diverge more significantly. This approach enables a thorough exploration of the local solution space initially, followed by the generation of more diverse hypotheses, thereby enhancing the robustness of the final pose estimation by covering a wider range of potential solutions. The final transformation is evaluated by:
\begin{equation}
    {\bf{\hat R,\hat t = }}\mathop {{\rm{argmax}}}\limits_{{{\bf{R}}_j},{{\bf{t}}_j}} \sum\limits_{i = 1}^{\left| {\bf{C}_{\text{initial}}} \right|} \phi  (\left\| {{{\bf{R}}_j}{\bf{p}}{}_{i}^s + {{\bf{t}}_j} - {\bf{p}}{}_{i}^t} \right\|),
\end{equation}
where $\phi$ is the Mean Absolute Error (MAE) function~\cite{yang2021toward}. 

\subsection{Loss Functions}
To supervise HyperGCT learning, a joint loss function is proposed:
\begin{equation}
\mathcal{L}={\mathcal{L}_{class}}+{\mathcal{L}_{match}}+{\mathcal{L}_{graph}},
\end{equation}
where ${\mathcal{L}_{class}}$ denotes the classification loss to supervise each correspondence, ${\mathcal{L}_{match}}$ denotes the matching loss to supervise pairs of correspondences, and ${\mathcal{L}_{graph}}$ denotes the graph loss to supervise the hypergraph. 

For the classification loss, HyperGCT uses binary cross-entropy loss to learn the confidence scores ${\bf{\hat s}}$ for ${\bf C}_{\text{initial}}$:
\begin{equation}
    {\mathcal{L}_{class}} = {{BCE({\bf{\hat s}},}}{{\bf{s}}^{*}}{\rm{)}},
\end{equation}
where ${\bf{s}}^{*}$ represents the ground truth labels of inlier/outlier:
    ${{\bf{s}}^*} = \left[ {\left\| {{{\bf{R}}^*}{\bf{p}}{}_i^s + {{\bf{t}}^*} - {\bf{p}}{}_i^t} \right\| < \theta_\text{inlier} } 
 \right]$.

For the match loss, a spectral matching loss~\cite{bai2021pointdsc} is used, which formulated as follows:
\begin{equation}
    {\mathcal{L}_{match}} = \frac{1}{{{{\left| {{{\bf C}_{\text{initial}}}} \right|}^2}}}\sum\limits_{ij} {{{({\eta _{ij}} - {\eta _{ij}^*})}^2}},
\end{equation}
where $\eta_{ij}$ is the compatibility score of the pairwise correspondence features:
    ${\eta_{ij}}=\left[1 - \frac{1}{{{\sigma _\text{f}}^2}}{\left\| {{{\bf{X}}_i} - {{\bf{X}}_j}} \right\|^2}\right]_+$,
where ${\sigma _\text{f}}$ is a learnable parameter that controls the sensitivity to feature differences. The term $\eta_{ij}^*$ is the ground-truth compatibility score between inliers.

The graph loss is formulated as follows:
\begin{equation}
    {\mathcal{L}_{graph}} = \frac{1}{{\left| {{{\bf C}_{\text{initial}}}} \right|}}\sum\limits_i {BCE({{\bf W}_H(i,:)},{{\bf H}^*(i,:)})},
\end{equation}
${\bf W}_H$ and ${\bf H}^*$ are the weight matrix output by HyperGCT and the ground truth incidence matrix formed by inliers.

%% file: sec/4_exp.tex
\section{Experiments}
\subsection{Experimental Setup}
\noindent\textbf{Dataset.} We use indoor scene datasets 3DMatch/3DLoMatch~\cite{zeng20173dmatch,huang2021predator}, as well as outdoor scene datasets KITTI-10m~\cite{qiao2024g3reg} and KITTI-LC~\cite{qiao2024g3reg}, for training and testing. Additionally, we consider the outdoor scene dataset ETH~\cite{pomerleau2012challenging} for generalization evaluation. For 3DMatch, following ~\cite{choy2020deep,bai2021pointdsc}, we use the provided train/validation/test split and evaluate on 1623 pairs from 8 scenes. For 3DLoMatch, following ~\cite{huang2021predator}, 1781 low overlap pairs are used for evaluation. For KITTI-10m, we follow the split in~\cite{choy2020deep,bai2021pointdsc}, with 0-5/6-7/8-10 sequences for train/val/test. For KITTI-LC, we conduct comparisons across 0-10m, 10-20m, and 20-30m translation ranges, comprising 914, 1151, and 1260 pairs, respectively. ETH contains 713 clutter and occlusion pairs from 4 scenes. Please refer to Supp.~\ref{sec:dataset} for more details.
\begin{table*}[t]
\centering
\caption{Registration results on 3DMatch. The key metric is \textbf{RR}, as it reflects the number of successfully registered pairs and overall performance. RE and TE focus on local precision and are meaningful for comparison only when RR is similar.}
\resizebox{\linewidth}{!}{
\begin{tabular}{l|ccccc>{\columncolor{gray!20}}cc|ccccc>{\columncolor{gray!20}}cc}
\bottomrule
         & \multicolumn{7}{c|}{FPFH~\cite{rusu2009fast}}                                                                                         & \multicolumn{7}{c}{FCGF~\cite{choy2019fully}}                                                                                          \\ \cline{2-15} 
         & IP ($\%$$\uparrow$)             & IR ($\%$$\uparrow$)            & {F1} ($\%$$\uparrow$)            & RE ($\degree$$\downarrow$)           & TE ($cm$$\downarrow$)            & \textbf{RR} ($\%$$\uparrow$)             & TIME ($s$)          & IP ($\%$$\uparrow$)             & IR ($\%$$\uparrow$)             & {F1} ($\%$$\uparrow$)             & RE ($\degree$$\downarrow$)           & TE ($cm$$\downarrow$)            & \textbf{RR} ($\%$$\uparrow$)             & TIME ($s$)         \\ \hline
SC$^2$-PCR~\cite{chen2023sc}   & 74.03          & 79.34          & 76.38          & 2.12          & 6.69          & 83.80          & \textbf{0.03} & 79.91          & 87.13          & 83.06          & 2.06          & 6.53          & 93.16          & \textbf{0.03} \\
MAC~\cite{yang2024mac}      & {\ul 74.56}          & 78.96          & 76.55          & 2.11          & 6.79          & {\ul 83.92}          &        1.77       & 79.93          & 86.64          & 82.84          & {\ul 2.03}          & 6.56          & {\ul 93.78}          &       1.80        \\ \hline
PointDSC~\cite{bai2021pointdsc} & 68.62          & 71.79          & 69.93          & \textbf{2.04} & \textbf{6.42} & 77.51          & {\ul 0.06}          & 78.97          & 86.25          & 82.17          & 2.06          & 6.57          & 93.22          & {\ul 0.06}          \\
PG-Net~\cite{wang2023pg}   & 71.57          & 76.60          & 73.82          & 2.30          & 6.73          & 82.32          & 0.19          & 79.25          & 86.53          & 82.43          & 2.15          & 6.51          & 93.10          & 0.19          \\
VBReg~\cite{jiang2023robust}    & 70.10          & 72.61          & 71.13          & 2.30          & 7.00          & 79.73          & 0.19          & 79.79          & 86.12          & 82.57          & {\ul 2.03}          & 6.47          & 92.98          & 0.19          \\
Hunter~\cite{yao2023hunter}   & 71.79          & 76.96          & 74.12          & 2.12          & 6.63          & 82.38          & 0.12          & 79.19          & 85.29          & 81.90          & 2.06          & 6.54          & 93.47          & 0.11          \\
3DPCP~\cite{wang20243dpcp}    & 74.32          & \textbf{80.83} & {\ul 76.64}          & {\ul 2.05}          & {\ul 6.50}          & 81.15          & 0.11          & \textbf{89.65} & \textbf{95.94} & \textbf{91.25} & \textbf{1.95} & \textbf{6.31} & 92.54          & 0.11          \\
HyperGCT & \textbf{75.15} & {\ul 80.10}          & \textbf{77.38} & 2.26          & 6.74          & \textbf{85.89} & 0.20          & {\ul 80.72}          & {\ul 87.56}          & {\ul 83.75}          & 2.04          & {\ul 6.34}          & \textbf{94.45} & 0.21          \\ \toprule
\end{tabular}}
\label{tab:3dmatch}
\end{table*}
\begin{table*}
\centering
\caption{Registration results on 3DLoMatch.}
\resizebox{\linewidth}{!}{
\begin{tabular}{l|ccccc>{\columncolor{gray!20}}cc|ccccc>{\columncolor{gray!20}}cc}
\bottomrule
                                & \multicolumn{7}{c|}{FPFH~\cite{rusu2009fast}}                                                                                                                                                    & \multicolumn{7}{c}{FCGF~\cite{choy2019fully}}                                                                                                                                \\ \cline{2-15} 
                                & IP ($\%$$\uparrow$) & IR ($\%$$\uparrow$) & {F1} ($\%$$\uparrow$) & RE ($\degree$$\downarrow$) & TE ($cm$$\downarrow$) & \textbf{RR} ($\%$$\uparrow$) & TIME ($s$)    & IP ($\%$$\uparrow$) & IR ($\%$$\uparrow$) & {F1} ($\%$$\uparrow$) & RE ($\degree$$\downarrow$) & TE ($cm$$\downarrow$) & \textbf{RR} ($\%$$\uparrow$) & TIME ($s$)    \\ \hline
SC$^2$-PCR~\cite{chen2023sc}    & 32.46               & 38.24               & 34.74                        & 4.04                       & 10.32                 & 38.46                        & \textbf{0.03} & 46.44               & 56.05               & 50.25                        & 3.79                       & {\ul 10.37}           & 58.62                        & \textbf{0.03} \\
MAC~\cite{yang2024mac}          & \textbf{35.27}      & \textbf{40.8}       & \textbf{37.45}               & 4.04                       & 10.61                 & {\ul 41.38}                  & 1.24          & 44.97               & 54.05               & 48.51                        & \textbf{3.75}              & 10.61                 & {\ul 60.08}                  & 1.26          \\ \hline
PointDSC~\cite{bai2021pointdsc} & 25.53               & 27.68               & 26.2                         & \textbf{3.81}              & \textbf{9.80}         & 29.20                        & {\ul 0.05}    & 44.35               & 52.47               & 47.57                        & {\ul 3.76}                 & 10.47                 & 57.33                        & {\ul 0.05}    \\
PG-Net~\cite{wang2023pg}        & 30.59               & 35.5                & 32.52                        & 5.05                       & 10.30                 & 37.11                        & 0.16          & 44.97               & 54.07               & 48.58                        & 4.42                       & 10.53                 & 58.39                        & 0.17          \\
VBReg~\cite{jiang2023robust}    & 28.46               & 31.2                & 29.41                        & 4.13                       & 10.61                 & 32.57                        & 0.20          & 45.24               & 53.44               & 48.48                        & 3.81                       & 10.68                 & 57.50                        & 0.20          \\
Hunter~\cite{yao2023hunter}     & 31.37               & 36.85               & 33.54                        & {\ul 3.96}                 & 10.15                 & 38.24                        & 0.09          & 46.48               & 55.11               & 49.97                        & 3.79                       & 10.51                 & 60.02                        & 0.09          \\
3DPCP~\cite{wang20243dpcp}      & 28.13               & 35.15               & 30.39                        & {\ul 3.96}                 & {\ul 9.99}            & 32.51                        & 0.10          & \textbf{49.35}      & \textbf{65.29}      & \textbf{53.47}               & \textbf{3.75}              & 10.44                 & 56.71                        & 0.10          \\
HyperGCT                        & {\ul 34.08}         & {\ul 39.57}         & {\ul 36.27}                  & 4.81                       & 10.8                  & \textbf{42.28}               & 0.18          & {\ul 48.99}         & {\ul 58.00}         & {\ul 52.59}                  & 4.18                       & \textbf{10.17}        & \textbf{63.73}               & 0.20          \\ \toprule
\end{tabular}
}
\label{tab:3dlomatch}
\end{table*}

\noindent\textbf{Evaluation Criteria.} 
For all experiments, we report registration recall (RR), which measures the rate of successful pairs whose rotation error (RE) and translation error (TE) fall below specific thresholds. Following~\cite{choy2020deep,bai2021pointdsc,yang2024mac,yang20243d}, successful registration is considered when RE$\leq$15$\degree$ and TE$\leq$30$cm$ for 3DMatch/3DLoMatch, and ETH, and RE$\leq$5$\degree$ and TE$\leq$60$cm$ for KITTI-10m. For the KITTI-LC, the subsets ``0-10m'', ``10-20m'', and ``20-30m'' use TE thresholds of 60$cm$, 120$cm$, and 180$cm$, respectively, with the same RE threshold as KITTI-10m. Inlier precision (IP), inlier recall (IR), and F1-score (F1) are reported to assess the outlier removal performance.

\noindent\textbf{Implementation Details.} Our method is implemented in PyTorch. For 3DMatch/3DLoMatch and KITTI-10m, we generate FCGF~\cite{choy2019fully} descriptors with the voxel size of 5$cm$ and 30$cm$, respectively, and randomly sample 1000 points with computed FCGF features to build correspondences for training. We set parameter $\sigma_\text{d}$ to 0.1$m$ for 3DMatch and 1.2$m$ for KITTI-10m, and initialize $\sigma_\text{f}$ to 1.0. Using a batch size of 6, we train the network for 50 epochs using the ADAM optimizer, starting with an initial learning rate of 0.0001 and a decay factor of 0.99. The number of seeds $N_\text{s}$ is set to 0.2$\left|{\bf C}_{\text{initial}}\right|$, and $N_{\text{initial}}$ is set to 0.1$N_\text{s}$. Following~\cite{tennakoon2015robust}, the size of the sampled minimal is set to 6, with a maximum of 30 iterations and a step size of 3. All experiments are conducted on a machine with an Intel Xeon E5-2690 CPU and NVIDIA RTX3090 GPUs. To fairly compare the actual standalone performance, we $\emph{DO NOT}$ apply ICP~\cite{segal2009generalized} to refine the results of all the methods.

\subsection{Results on Indoor Scenes}\label{sec:indoor}
We compare our method with two traditional methods SC$^2$-PCR~\cite{chen2023sc}, MAC~\cite{yang2024mac}, and five deep-learned outlier removal methods PointDSC~\cite{bai2021pointdsc}, PG-Net~\cite{wang2023pg}, VBReg~\cite{jiang2023robust}, Hunter~\cite{yao2023hunter}, 3DPCP~\cite{wang20243dpcp}. Note that we retrain all deep-learned methods for testing. We generate correspondences using FPFH~\cite{rusu2009fast} and FCGF~\cite{choy2019fully} descriptors following~\cite{bai2021pointdsc,chen2023sc}. As shown in Table~\ref{tab:3dmatch}, HyperGCT achieves the highest RR of 85.89$\%$ with FPFH and 94.45$\%$ with FCGF, outperforming all the compared methods. Notably, when combined with FPFH, our method surpasses 3DPCP, a state-of-the-art correspondence pruning network, in terms of IP and F1. As shown in Table~\ref{tab:3dlomatch}, HyperGCT also outperforms the others, achieving RR of 42.28$\%$ with FPFH and 63.73$\%$ with FCGF. Additionally, our method has the same time consumption as VBReg. HyperGCT is trained using only FCGF correspondences, yet it still achieves the best performance when combined with other descriptors, demonstrating the strong generalization ability of our approach. Please refer to Supp.~\ref{sec:more_result} for results with other methods.

\subsection{Results on Outdoor Scenes}
\begin{table}[t]
\centering
\caption{Registration results (RR) on KITTI-10m.}
\resizebox{\linewidth}{!}{
\begin{tabular}{l|cccc|cccc}
\bottomrule
           & \multicolumn{4}{c|}{FPFH~\cite{rusu2009fast}}                                                                                   & \multicolumn{4}{c}{FCGF~\cite{choy2019fully}}                                                                                   \\ \cline{2-9} 
           & \multicolumn{1}{c}{8} & \multicolumn{1}{c}{9} & \multicolumn{1}{c}{10} & \multicolumn{1}{c|}{\textbf{AVG.}} & \multicolumn{1}{c}{8} & \multicolumn{1}{c}{9} & \multicolumn{1}{c}{10} & \multicolumn{1}{c}{\textbf{AVG.}} \\ \hline
SC$^2$-PCR~\cite{chen2023sc} &           98.37            &  \textbf{99.38}                     &     \textbf{100}                   &              {\ul 98.92}                      &                {\ul 99.02}       &     {\ul 97.52}                  &              {\ul 98.85}          &       98.56                            \\
MAC~\cite{yang2024mac}        &           \textbf{99.35}            &   {\ul 98.76}                    &         {\ul 98.84}               &              \textbf{99.10}                      &                 98.37      &     \textbf{98.14}                  &             {\ul 98.85}           &      98.38                             \\ \hline
PointDSC~\cite{bai2021pointdsc}   &           97.71            &    98.14                   &         \textbf{100}               &           98.19                         &      {\ul 99.02}                 &          {\ul 97.52}             &    97.70                    &       98.38                            \\
PG-Net~\cite{wang2023pg}     &           97.71            &     {\ul 98.76}                  &        \textbf{100}                &            98.38                        &          {\ul 99.02}             &            {\ul 97.52}           &       {\ul 98.85}                 &       98.56                            \\
VBReg~\cite{jiang2023robust}      &          98.37             &  98.14                     &        \textbf{100}                &                    98.56                &         {\ul 99.02}              &                \textbf{98.14}       &           {\ul 98.85}             &      {\ul 98.74}                             \\
Hunter~\cite{yao2023hunter}     &             98.37          &      {\ul 98.76}                 &     \textbf{100}                   &                {98.74}                    &              {\ul 99.02}         &        {\ul 97.52}               &             \textbf{100}           &      {\ul 98.74}                             \\
3DPCP~\cite{wang20243dpcp}      &         {\ul 98.69}              &  \textbf{99.38}                     &       \textbf{100}                 &               \textbf{99.10}                     &          {\ul 99.02}             &         95.65              &       97.70                 &      97.83                             \\
HyperGCT   &         {\ul 98.69}              &   \textbf{99.38}                    &      \textbf{100}                  &                \textbf{99.10}                    &       \textbf{99.35}                &           {\ul 97.52}            &     \textbf{100}                   &    \textbf{98.92}                               \\ \toprule
\end{tabular}}
\label{tab:kitti-10m}
\end{table}
\begin{table}[t]
    \centering
    \caption{Registration results (RR) on KITTI-LC with FPFH.}
    \resizebox{0.6\linewidth}{!}{
    \begin{tabular}{l|c|c|c}
\bottomrule
                                & 0-10m         & 10-20m         & 20-30m         \\ \hline
SC$^2$-PCR~\cite{chen2023sc}    & 95.84         & 61.86          & 18.25          \\
MAC~\cite{yang2024mac}          & \textbf{97.7} & {\ul 69.24}    & {\ul 23.73} \\ \hline
PointDSC~\cite{bai2021pointdsc} & 93.33         & 53.87          & 17.54          \\
PG-Net~\cite{wang2023pg}        & 93.00         & 42.75          & 10.48          \\
VBReg~\cite{jiang2023robust}    & 96.28         & 56.91          & 17.46          \\
Hunter~\cite{yao2023hunter}     & 92.34         & 50.22          & 13.89          \\
3DPCP~\cite{wang20243dpcp}      & 96.94   & 67.25          & 23.49    \\
HyperGCT                       & {\ul 97.05}   & {\bf 72.46} & {\bf 25.95}     
\\ \toprule
\end{tabular}}
\label{tab:kitti-lc}
\end{table}

Following~\cite{qiao2024g3reg}, we conduct comparisons on KITTI-10m and KITTI-LC for a comprehensive evaluation, using the same methods as in the indoor scene experiments. As shown in Table~\ref{tab:kitti-10m}, our method achieves the highest RR of 99.10$\%$ with FPFH and 98.92$\%$ with FCGF on KITTI-10m, with a more pronounced performance gap using FCGF. On KITTI-LC, as the translation distance increases, the outlier ratio rises, posing a challenge for registration. As shown in Table~\ref{tab:kitti-lc}, HyperGCT outperforms other deep-learned methods on the ``0-10m'' subset and surpasses all others on the ``10-20m'' and ``20-30m'' subset, demonstrating its flexibility in handling correspondences of varying quality.

\subsection{Generalization Results}
We conduct generalization experiments across different data modalities (3DMatch, KITTI-10m, and ETH). Methods for comparison include SC$^2$-PCR~\cite{chen2023sc}, MAC~\cite{yang2024mac}, PointDSC~\cite{bai2021pointdsc}, PG-Net~\cite{wang2023pg}, VBReg~\cite{jiang2023robust}, Hunter~\cite{yao2023hunter}, and 3DPCP~\cite{wang20243dpcp}. No parameter tuning is performed.

\noindent\textbf{Between 3DMatch and KITTI-10m.}
\begin{table}[t]
\centering
\caption{Generalize between 3DMatch and KITTI-10m.}
\resizebox{0.85\linewidth}{!}{
\begin{tabular}{l|cc|cc}
\bottomrule
         & \multicolumn{2}{c|}{3DMatch$\rightarrow$KITTI-10m} & \multicolumn{2}{c}{KITTI-10m$\rightarrow$3DMatch} \\ \cline{2-5} 
         & FPFH~\cite{rusu2009fast}             & FCGF~\cite{choy2019fully}             & FPFH~\cite{rusu2009fast}             & FCGF~\cite{choy2019fully}            \\ \hline
SC$^2$-PCR~\cite{chen2023sc} &           {\ul 98.92}           &  {\ul 98.56}                     &     83.80                   &              93.16                                               \\
MAC~\cite{yang2024mac}        &           \textbf{99.10}            &   98.38                    &         {\ul 83.92}               &          {\ul 93.78}                                                   \\
PointDSC~\cite{bai2021pointdsc} & 91.16            & {\ul 98.56}      & 72.46            & 92.42           \\
PG-Net~\cite{wang2023pg}   & 50.36            & 94.22            & 76.22            & 91.99           \\
VBReg~\cite{jiang2023robust}    & 97.11            & {\ul 98.56}   & 76.16            & 92.17           \\
Hunter~\cite{yao2023hunter}   & {98.01}      & \textbf{98.92}            & {82.13}      & {93.47}     \\
3DPCP~\cite{wang20243dpcp}    & {98.01}            &   98.01          & 54.71            & 83.43           \\
HyperGCT & \textbf{99.10}   & \textbf{98.92}            & \textbf{85.40}   & \textbf{94.39}  \\ \toprule
\end{tabular}}
\label{tab:3dmatch-kitti}
\end{table}
The direct generalization performance of the five comparison methods is lower than their results when trained on the corresponding datasets. In contrast, our method surpasses all comparison methods and maintains strong generalization capability. As shown in Table~\ref{tab:3dmatch-kitti}, our method's generalization performance on KITTI is consistent with the performance when trained on KITTI, and its generalization performance on 3DMatch is slightly lower ($<$0.5$\%$) than when trained on 3DMatch.

\noindent\textbf{From 3DMatch, KITTI-10m to ETH.}
\begin{table}[t]
\centering
\caption{Generalize from 3DMatch to ETH.}
\resizebox{0.85\linewidth}{!}{
\begin{tabular}{l|ccccc}
\bottomrule
         & \multicolumn{5}{c}{FPFH~\cite{rusu2009fast}}                                                                                                                             \\ \cline{2-6} 
         & \multicolumn{1}{c}{summer} & \multicolumn{1}{c}{winter} & \multicolumn{1}{c}{autmn} & \multicolumn{1}{c}{summer} & \multicolumn{1}{c}{\textbf{AVG.}} \\ \hline
SC$^2$-PCR~\cite{chen2023sc}  & \textbf{57.07} & \textbf{34.60} & {\ul 46.96} & {\ul 54.40} & {\ul 45.06}                \\
 MAC~\cite{yang2024mac} & 46.74 & 27.68 & 33.04 & 43.20 & 36.12 \\
PointDSC~\cite{bai2021pointdsc} & 25.00                      & 14.19                      & 3.48                      & 9.60                       & 14.45                             \\
PG-Net~\cite{wang2023pg}   & 41.85                      & 21.80                      & 24.35                     & 32.80                      & 29.31                             \\
VBReg~\cite{jiang2023robust}    & 38.59                      & 22.15                      & 16.52                     & 24.80                      & 25.95                             \\
Hunter~\cite{yao2023hunter}   & 47.28                      & 25.95                      & {42.61}               & 44.00                      & 37.31                             \\
3DPCP~\cite{wang20243dpcp}    & {\ul 53.26}                & {\ul 32.87}             & 41.74                     & {44.80}                & {41.65}                       \\
HyperGCT & \textbf{57.07}             & { 31.83}                & \textbf{50.43}            & \textbf{56.80}             & \textbf{45.72}                    \\ \hline
         & \multicolumn{5}{c}{FCGF~\cite{choy2019fully}}                                                                                                                             \\ \cline{2-6} 
         & \multicolumn{1}{c}{summer} & \multicolumn{1}{c}{winter} & \multicolumn{1}{c}{autmn} & \multicolumn{1}{c}{summer} & \multicolumn{1}{c}{\textbf{AVG.}} \\ \hline
SC$^2$-PCR~\cite{chen2023sc}  & \textbf{79.89} & {\ul 49.83} & 78.26 & {\ul 79.20} & \textbf{67.32}   \\
 MAC~\cite{yang2024mac} & 75.54 & 42.91 & 71.30 & 73.60 & 61.29 \\
PointDSC~\cite{bai2021pointdsc} & 41.85                      & 26.99                      & 33.91                     & 45.60                       & 35.2                              \\
PG-Net~\cite{wang2023pg}   & {71.20}                       & 39.10                       & 64.35                     & 68.00                         & 56.52                             \\
VBReg~\cite{jiang2023robust}    & 66.30                       & 35.29                      & 66.09                     & 59.20                       & 52.45                             \\
Hunter~\cite{yao2023hunter}   & 61.96                      & 27.34                      & 61.74                     & 61.60                       & 50.77                             \\
3DPCP~\cite{wang20243dpcp}    & {\ul 79.35}                & \textbf{51.56}             & {\ul 79.13}            & {69.60}                       & {66.34}                    \\
HyperGCT & {\ul 79.35}             & {47.06}                 & \textbf{80.87}               & \textbf{80.00}              & {\ul 66.62}                       \\ \toprule
\end{tabular}}
\label{tab:3dmatch-eth}
\end{table}
We further test the generalization of all methods on ETH, which represents more complex geometries and challenges such as clutter and occlusion. As shown in Tables~\ref{tab:3dmatch-eth} and \ref{tab:kitti-eth}, our method outperforms the others in all settings. Moreover, the results of our method, generalized from 3DMatch and KITTI-10m are relatively consistent, indicating strong generalization capability across different datasets.

\begin{table}[t]
\centering
\caption{Generalize from KITTI-10m to ETH.}
\resizebox{0.85\linewidth}{!}{
\begin{tabular}{l|ccccc}
\bottomrule
         & \multicolumn{5}{c}{FPFH~\cite{rusu2009fast}}                                                                                                                             \\ \cline{2-6} 
         & \multicolumn{1}{c}{summer} & \multicolumn{1}{c}{winter} & \multicolumn{1}{c}{autmn} & \multicolumn{1}{c}{summer} & \multicolumn{1}{c}{\textbf{AVG.}} \\ \hline
SC$^2$-PCR~\cite{chen2023sc}  & {\ul 57.07} & 34.60 & {\ul 46.96} & \textbf{54.40} & {\ul 45.06}                \\
 MAC~\cite{yang2024mac} & 46.74 & 27.68 & 33.04 & 43.20 & 36.12 \\
PointDSC~\cite{bai2021pointdsc} & 17.93                      & 8.30                       & 6.96                      & 5.60                       & 10.10                             \\
PG-Net~\cite{wang2023pg}   & 22.28                      & 10.38                      & 6.96                      & 8.00                       & 12.48                             \\
VBReg~\cite{jiang2023robust}    & 41.85                      & 26.30                      & 30.43                     & 32.80                      & 32.12                             \\
Hunter~\cite{yao2023hunter}   & {54.89}                & 26.64                      & { 42.61}               & {45.60}                & 39.83                             \\
3DPCP~\cite{wang20243dpcp}    & 53.26                      & \textbf{37.37}             & 36.52                     & 40.80                      & {41.94}                       \\
HyperGCT & \textbf{57.61}             & {\ul 35.99}                & \textbf{49.57}            & {\ul 51.20}             & \textbf{46.42}                    \\ \hline
         & \multicolumn{5}{c}{FCGF~\cite{choy2019fully}}                                                                                                                             \\ \cline{2-6} 
         & \multicolumn{1}{c}{summer} & \multicolumn{1}{c}{winter} & \multicolumn{1}{c}{autmn} & \multicolumn{1}{c}{summer} & \multicolumn{1}{c}{\textbf{AVG.}} \\ \hline
SC$^2$-PCR~\cite{chen2023sc}  & {\ul 79.89} & \textbf{49.83} & {\ul 78.26} & {\ul 79.20} & {\ul 67.32}   \\
 MAC~\cite{yang2024mac} & 75.54 & 42.91 & 71.30 & 73.60 & 61.29 \\
PointDSC~\cite{bai2021pointdsc} & 41.85                      & 23.18                      & 36.52                     & 36.80                      & 32.54                             \\
PG-Net~\cite{wang2023pg}   & 52.17                      & 30.80                      & 44.35                     & 36.80                      & 39.55                             \\
VBReg~\cite{jiang2023robust}    & 62.50                      & 45.33                      & {72.17}            & {68.00}                & 58.06                             \\
Hunter~\cite{yao2023hunter}   & {67.39}                & 30.45                      & 63.48                     & 61.60                      & 50.77                             \\
3DPCP~\cite{wang20243dpcp}    & 66.30                      & {47.06}             & {70.43}               & 60.80                      & {58.20}                       \\
HyperGCT & \textbf{81.52}             & {\ul 47.75}                & \textbf{80.00}            & \textbf{80.80}             & \textbf{67.46}                    \\ \toprule
\end{tabular}}
\label{tab:kitti-eth}
\end{table}

\subsection{Analysis Experiments}
\noindent\textbf{Improvement of Graph Quality.}
\begin{figure}[t]
  \centering
   \includegraphics[width=\linewidth]{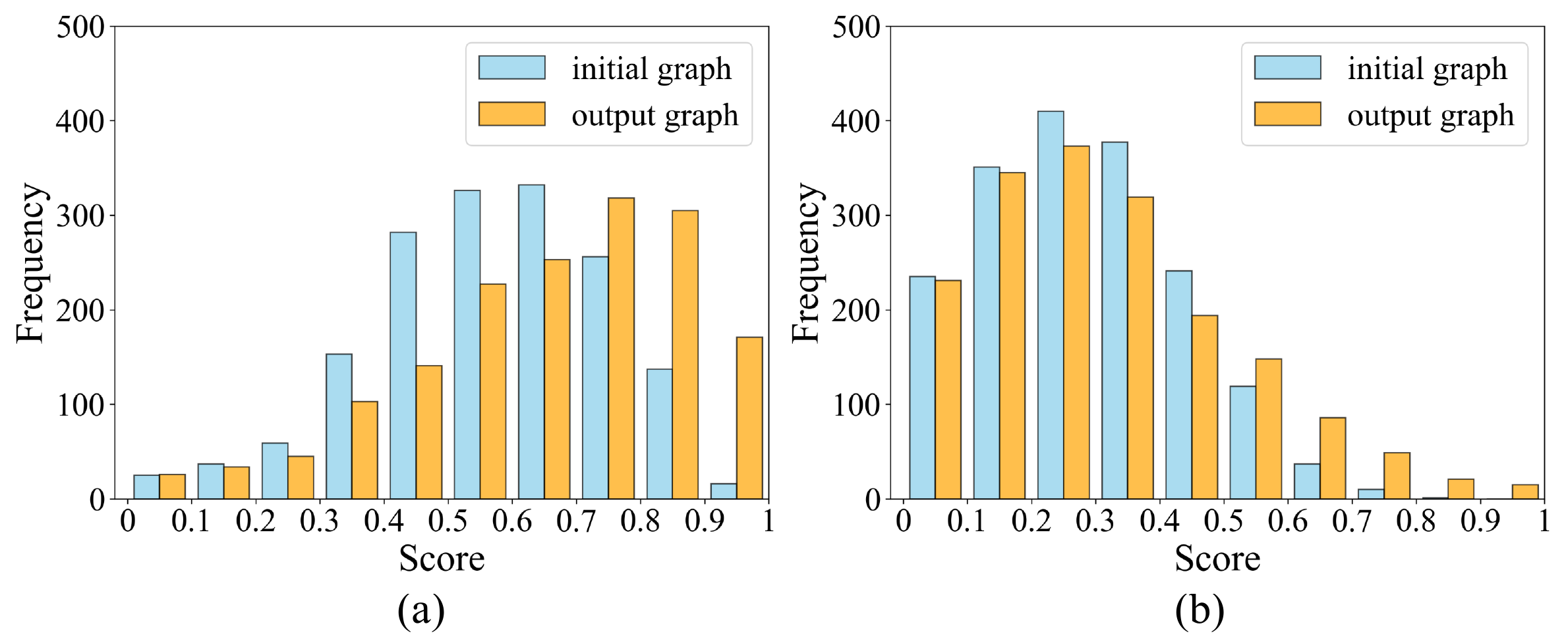}
   \caption{Improvement of graph quality HyperGCT brings when combined with FCGF on 3DMatch (a) and 3DLoMatch (b).}
   \label{fig:improve}
\end{figure}
To assess the graph quality improvement brought by HyperGCT, we introduce a metric called hyperedge precision $\mathcal{P}(\pmb{\mathcal{E}})$, defined as:
\begin{equation}
    \mathcal{P}({\pmb{\mathcal{E}}}) = \frac{1}{\left|\pmb{\mathcal{E}}\right|}\sum\limits_{i} {\frac{{\left| {{{e}}_i^{\text{correct}}} \right|}}{{\left| {{{{e}}_i}} \right|}}}, 
\end{equation}
where ${{e}}_i^{\text{correct}} = \{ {\rm{v}}_i^j|{\rm{v}}_i^j \in {{{e}}_i} \cap {{e}}_i^{\text{inlier}}\}$ and ${{e}}_i^{\text{inlier}}$ denotes the ground truth hyperedge formed by inlier labels. Scores of $\mathcal{P}({\pmb{\mathcal{E}}})$ lie in [0,1] and are divided into 10 equal bins for frequency analysis on 3DMatch and 3DLoMatch when combined with FCGF. As shown in Fig.~\ref{fig:improve}, the frequency in the lower $\mathcal{P}({\pmb{\mathcal{E}}})$ range decreases in the hypergraph output by HyperGCT, while the frequency in the higher $\mathcal{P}({\pmb{\mathcal{E}}})$ range increases. This indicates that HyperGCT effectively captures the high-order consistency of inliers around the nodes, thereby improving the overall hyperedge precision.

\noindent\textbf{Robustness to Noise in Graph Construction.}
\begin{table}[t]
\centering
\caption{Robustness to threshold $\theta_\text{cmp}$. `\ding{55}' denotes out of memory. $\downarrow$ / $\uparrow$ denote the decline / rise rate ($\%$) compared to default. }
\resizebox{\linewidth}{!}{
\begin{tabular}{ccc|c|ccccc}
\bottomrule
                                                &                                            &          & Default & 0.8 & 0.85 & 0.9 & 0.95 & 0.99 \\ \hline
\multicolumn{1}{c|}{\multirow{4}{*}{3DMatch}}   & \multicolumn{1}{c|}{\multirow{2}{*}{FPFH}} & MAC      & 83.92   & \ding{55}    &   \ding{55}   &   \ding{55}  &   \ding{55}   &   $\downarrow$0.49   \\
\multicolumn{1}{l|}{}                           & \multicolumn{1}{c|}{}                      & HyperGCT & 85.89   &  $\downarrow$0.06   &  $\downarrow$0.37    &  $\downarrow$0.25   &  $\downarrow$0.25    & $\downarrow$0.68     \\ \cline{2-9}
\multicolumn{1}{l|}{}                           & \multicolumn{1}{c|}{\multirow{2}{*}{FCGF}} & MAC      & 93.78   &  \ding{55}   &   \ding{55}  &   $\downarrow$2.90  &  $\downarrow$2.10    &   $\downarrow$0.31   \\
\multicolumn{1}{l|}{}                           & \multicolumn{1}{c|}{}                      & HyperGCT & 94.45   &  $\downarrow$0.06   &   $\downarrow$0.06   &  $\downarrow$0.12   &  $\downarrow$0.12    &   $\downarrow$0.06   \\ \hline
\multicolumn{1}{c|}{\multirow{4}{*}{3DLoMatch}} & \multicolumn{1}{l|}{\multirow{2}{*}{FPFH}} & MAC      & 41.38   &   \ding{55}  &   \ding{55}   &  $\downarrow$0.05   &   $\downarrow$0.45   &   0   \\
\multicolumn{1}{c|}{}                           & \multicolumn{1}{l|}{}                      & HyperGCT & 42.28   &  $\uparrow$0.50 & $\uparrow$0.79     & $\uparrow$0.73    & $\uparrow$0.67     & $\uparrow$0.17     \\ \cline{2-9}
\multicolumn{1}{c|}{}                           & \multicolumn{1}{l|}{\multirow{2}{*}{FCGF}} & MAC      & 60.08   &  $\downarrow$5.84   &   $\downarrow$5.50   &  $\downarrow$3.82   &   $\downarrow$0.73   &   0   \\
\multicolumn{1}{c|}{}                           & \multicolumn{1}{l|}{}                      & HyperGCT & 63.73   &  $\uparrow$0.05   &  $\uparrow$0.34    &  0   &   $\downarrow$0.11   & $\uparrow$0.05     \\ \toprule
\end{tabular}}
\label{tab:measurement_error}
\end{table}
\begin{table}[t]
\centering
\caption{Robustness to threshold $\theta_\text{inlier}$.}
\resizebox{0.9\linewidth}{!}{
\begin{tabular}{ccc|c|ccc}
\bottomrule
\multicolumn{1}{l}{}                            & \multicolumn{1}{l}{}                       &          & 0.1 (Default) & 0.05 & 0.15 & 0.2 \\ \hline
\multicolumn{1}{c|}{\multirow{4}{*}{3DMatch}}   & \multicolumn{1}{c|}{\multirow{2}{*}{FPFH}} & MAC      & 83.92        &   $\downarrow$3.27   &   0   &   $\downarrow$0.06  \\
\multicolumn{1}{c|}{}                           & \multicolumn{1}{c|}{}                      & HyperGCT & 85.89        &   $\downarrow$0.68   &  $\downarrow$0.49    & $\downarrow$0.92    \\ \cline{2-7} 
\multicolumn{1}{c|}{}                           & \multicolumn{1}{c|}{\multirow{2}{*}{FCGF}} & MAC      & 93.78        &   $\downarrow$1.05   & $\downarrow$0.43     & $\downarrow$0.93    \\
\multicolumn{1}{c|}{}                           & \multicolumn{1}{c|}{}                      & HyperGCT & 94.45        &  $\downarrow$0.18    &   $\downarrow$0.12   & $\downarrow$0.12    \\ \hline
\multicolumn{1}{c|}{\multirow{4}{*}{3DLoMatch}} & \multicolumn{1}{c|}{\multirow{2}{*}{FPFH}} & MAC      & 41.38        &   $\downarrow$3.70   &  $\downarrow$1.85    & $\downarrow$3.76    \\
\multicolumn{1}{c|}{}                           & \multicolumn{1}{c|}{}                      & HyperGCT & 42.28        &  $\uparrow$0.79    &  $\downarrow$0.45    & $\downarrow$0.39    \\ \cline{2-7} 
\multicolumn{1}{c|}{}                           & \multicolumn{1}{c|}{\multirow{2}{*}{FCGF}} & MAC      & 60.08        &   $\downarrow$1.91   & $\downarrow$0.90     & $\downarrow$2.98    \\
\multicolumn{1}{c|}{}                           & \multicolumn{1}{c|}{}                      & HyperGCT & 63.73        &   $\downarrow$0.11   &  $\uparrow$0.22    & 0    \\ \toprule
\end{tabular}}
\label{tab:matching_error}
\end{table}
We compare our method with an explicit graph construction method, MAC~\cite{yang2024mac}, to test robustness against graph noise from measurement and matching errors. Measurement error affects correspondence compatibility (\ie, whether the compatibility score is less than threshold $\theta_\text{cmp}$), while matching error identifies inliers (\ie, whether the residual is below threshold $\theta_\text{inlier}$). As shown in Table~\ref{tab:measurement_error}, reducing $\theta_\text{cmp}$ causes MAC to face out-of-memory issues due to graph density, with performance dropping by up to 5.84$\%$ on 3DLoMatch using FCGF. In contrast, our method remains stable, with performance variation under 1$\%$. Similar trends with $\theta_\text{inlier}$ changes (Table~\ref{tab:matching_error}) highlight the superior robustness of our method to graph noise.

\noindent\textbf{Impact of Graph Update Frequency.} The frequency significantly influences performance by affecting both computational efficiency and the quality of information propagation within the network. As shown in Table~\ref{tab:graph_update}, excessive updates (more than 5 times) can increase model parameter sizes, thereby slowing convergence rates with the same training epochs. In contrast, insufficient updates (less than 5 times) allow more outliers to interfere with the information flow, which degrades overall performance. This highlights that there exists an optimal frequency for graph updates where computational resources are efficiently utilized and high-order relationships are accurately captured. 
\begin{table}[h]
\caption{Performance varying graph update times.}
\centering
\resizebox{\linewidth}{!}{
\begin{tabular}{c|c|cc|cc}
\bottomrule
\multicolumn{1}{l|}{\multirow{2}{*}{\# update}} & \multicolumn{1}{l|}{\multirow{2}{*}{\# params (M)}} & \multicolumn{2}{c|}{3DMatch}    & \multicolumn{2}{c}{3DLoMatch}   \\ \cline{3-6} 
\multicolumn{1}{l|}{}                           & \multicolumn{1}{l|}{}                               & FCGF (\%)      & FPFH (\%)      & FCGF (\%)      & FPFH (\%)      \\ \hline
1                                               & 0.30                                                & 93.96          & 84.47          & 63.05          & 40.93          \\
3                                               & 0.64                                                & 94.15 ($\uparrow$0.19)         & 85.15 ($\uparrow$0.68)         & 63.28 ($\uparrow$0.23)         & 41.66 ($\uparrow$0.73)         \\
5                                               & 0.98                                                & \textbf{94.45} ($\uparrow$0.49) & \textbf{85.89} ($\uparrow$1.42) & \textbf{63.73} ($\uparrow$0.68) & {42.28} ($\uparrow$1.35)       \\
7                                               & 1.32                                                & \textbf{94.45} ($\uparrow$0.49) & 84.72 ($\uparrow$0.25)          & 63.39 ($\uparrow$0.34)          & \textbf{42.56} ($\uparrow$1.63)\\
9                                               & 1.66                                                & 94.15 ($\uparrow$0.19)           & 85.09 ($\uparrow$0.62)         & 63.17 ($\uparrow$0.12)         & 41.61 ($\uparrow$0.68)         \\ \toprule
\end{tabular}}
\label{tab:graph_update}
\end{table}

\noindent\textbf{Combined with Different PCR Back-ends.}
\begin{table}[t]
\centering
\caption{Results of HyperGCT combined with other back-ends.}
\resizebox{0.85\linewidth}{!}{
\begin{tabular}{lcl|cc}
\bottomrule
                                                   & \multicolumn{1}{l}{}                            &      & SC$^2$-PCR & PLE   \\ \hline
\multicolumn{1}{l|}{\multirow{4}{*}{w/o HyperGCT}} & \multicolumn{1}{c|}{\multirow{2}{*}{3DMatch}}   & FPFH & 83.80  & 82.38 \\
\multicolumn{1}{l|}{}                              & \multicolumn{1}{c|}{}                           & FCGF & 93.16  & 93.47 \\ \cline{2-5}
\multicolumn{1}{l|}{}                              & \multicolumn{1}{c|}{\multirow{2}{*}{3DLoMatch}} & FPFH & 38.46  & 38.24 \\
\multicolumn{1}{l|}{}                              & \multicolumn{1}{c|}{}                           & FCGF & 58.62  & 60.02 \\ \hline
\multicolumn{1}{c|}{\multirow{4}{*}{w/ HyperGCT}}  & \multicolumn{1}{c|}{\multirow{2}{*}{3DMatch}}   & FPFH & 86.14 ($\uparrow$2.34)  & 82.99 ($\uparrow$0.61) \\
\multicolumn{1}{c|}{}                              & \multicolumn{1}{c|}{}                           & FCGF & 93.96 ($\uparrow$0.80)  & 93.90 ($\uparrow$0.43) \\ \cline{2-5}
\multicolumn{1}{c|}{}                              & \multicolumn{1}{c|}{\multirow{2}{*}{3DLoMatch}} & FPFH & 42.22 ($\uparrow$3.76)  & 38.97 ($\uparrow$0.73) \\
\multicolumn{1}{c|}{}                              & \multicolumn{1}{c|}{}                           & FCGF & 63.17 ($\uparrow$4.55)  & 60.08 ($\uparrow$0.06) \\ \toprule
\end{tabular}}
\label{tab:backend}
\end{table}
{
We integrate HyperGCT with other registration back-ends, such as SC$^2$-PCR and Hunter's PLE. This integration incorporates the learned incidence matrix $\bf H$ into their compatibility calculations. As shown in Table~\ref{tab:backend}, this integration significantly improves the performance of both registration back-ends. SC$^2$-PCR, in particular, demonstrates a more significant enhancement: for instance, a 2.34\% increase on the 3DMatch with FPFH and a 4.55\% increase on the 3DLoMatch with FCGF. The flexible constraint effectively enhances different solvers, demonstrating the adaptability and flexibility.}

\noindent\textbf{Correctness of Hypothesis Generation.}
\begin{table}[t]
\centering
\caption{Correctness ($\%$) of generated hypotheses.}
\resizebox{\linewidth}{!}{
\begin{tabular}{cl|cccc}
\bottomrule
\multicolumn{1}{l}{}                            &      & RANSAC~\cite{fischler1981random} & SAC-COT~\cite{yang2021sac} & MAC~\cite{yang2024mac}   & HyperGCT       \\ \hline
\multicolumn{1}{c|}{\multirow{2}{*}{3DMatch}}   & FPFH & 0.74       & 0.46    & 6.64  & \textbf{62.66} \\
\multicolumn{1}{c|}{}                           & FCGF & 10.17      & 9.35    & 23.62 & \textbf{87.62} \\ \hline
\multicolumn{1}{c|}{\multirow{2}{*}{3DLoMatch}} & FPFH & 0.04       & 0.02    & 0.93  & \textbf{14.61} \\
\multicolumn{1}{c|}{}                           & FCGF & 1.18       & 1.92    & 6.43  & \textbf{40.51} \\ \toprule
\end{tabular}}
\label{tab:hypo_precise}
\end{table}
To evaluate HyperGCT against random sampling, triplet, and maximal clique constraints, we assess its ability to generate valid hypotheses. As shown in Table~\ref{tab:hypo_precise}, HyperGCT demonstrates superior capability in producing accurate high-order consistency hypotheses compared to these traditional approaches,  which validates its robustness and effectiveness. Furthermore, the robustness of HyperGCT's solver enhances both the stability of the hypothesis generation process and the accuracy of pose estimation and verification.

%% file: sec/5_con.tex
\section{Conclusion}
This paper presents HyperGCT, a Hyper-GNN-learned geometric constraint for accurate hypothesis generation. HyperGCT is superior to competitors on several benchmarks, is robust to graph noise, and generalizes well across datasets. 
The GF-NMS module may not retrieve enough correct matches for accurate hypothesis generation. We will design a seeding module to select dynamically and reliably. \\
\noindent\textbf{Acknowledgments.} This work is partly supported by National Natural Science Foundation of China (No. 62372377 and 52172380) and China Postdoctoral Science Foundation (No. 2024M761014).

%% file: sec/X_suppl.tex
\clearpage
\setcounter{page}{1}
\maketitlesupplementary
\section{More Method Details}\label{sec:GF_NMS}
\noindent\textbf{GF-NMS.} First, we transform the incidence matrix $\bf H$ output by HyperGCT into an adjacent matrix $\bf A$ as follows:
\begin{equation}
    {\bf{A}} = {\left[ {{\bf{H}} + {{{\bf{H}}}^\top} - 1} \right]_ + }.
\end{equation}
The motivation is to check the consistency between vertices. If $e_i$ contains $v_j$ and $e_j$ contains $v_i$ (\ie, ${\bf H}(v_i,e_j)={\bf H}(v_j,e_i)=1$), then $v_i$ and $v_j$ are considered consistent, with ${\bf A}(v_i,v_j)={\bf A}(v_j,v_i)=1$.
Second, we apply a graph filter based on the degree signal of $\bf{A}$ to compute scores for each correspondence:
\begin{equation}
    {{{\bf s}}^{\text{GF}}} = {\rm{MinMax}}( {({{\bf D}_{\bf A}} - {\bf A})}{{\bf D}({\bf A})}),
\end{equation}
where ${\bf D}_{\bf A}$ is the diagonal degree matrix of ${\bf A}$ and ${{\bf D}({\bf A})}$ is the degree vector of ${\bf A}$. $\rm{MinMax}$ is the min-max normalization operation. 
 Higher values of ${{{\bf s}}^{\text{GF}}_i}$ indicate that $v_i$ has strong local connectivity, reflecting its structural importance within the graph. Third, we employ standard NMS to select correspondences with confidence scores $\hat{\bf s}$ that are local maxima, resulting in $N_1$ seeds ($N_1 \ll N_s$). The remaining correspondences are then ranked by ${{{\bf s}}^{\text{GF}}}$ in descending order, and the top $N_s-N_1$ ones are selected as the other part of the seed set. As shown in Table~\ref{tab:seeds}, GF-NMS generates twice as many inliers as NMS, indicating its effectiveness.
\begin{table}[h]
\centering
\caption{The average number of inliers among seeds.}
\resizebox{0.6\linewidth}{!}{
\begin{tabular}{cl|cc}
\bottomrule
\multicolumn{1}{l}{}                            &      & NMS    & GF-NMS          \\ \hline
\multicolumn{1}{c|}{\multirow{2}{*}{3DMatch}}   & FPFH & 64     & \textbf{152.65} \\
\multicolumn{1}{c|}{}                           & FCGF & 232.57 & \textbf{425.96} \\ \hline
\multicolumn{1}{c|}{\multirow{2}{*}{3DLoMatch}} & FPFH & 11.44  & \textbf{28.27}  \\
\multicolumn{1}{c|}{}                           & FCGF & 49.69  & \textbf{96.40}  \\ \toprule
\end{tabular}}
\label{tab:seeds}
\end{table}

\section{More Analysis.}\label{sec:more_result}
\noindent\textbf{Using Different Order Graphs.} We replace the second-order graph (SOG) with the first-order graph (FOG) to construct the initial hypergraph in Sect.~\ref{sec:graph_con} and conduct experiments on 3DMatch/3DLoMatch, KITTI-10m, and KITTI-LC. The results from Tables~\ref{tab:graph_order_1}, \ref{tab:graph_order_2} and \ref{tab:graph_order_3} show minimal performance differences between different order graphs, indicating that HyperGCT is robust and not sensitive to graph orders.
\begin{table}[h]
\caption{Registration results on 3DMatch/3DLoMatch.}
\centering
\resizebox{\linewidth}{!}{
\begin{tabular}{c|c|ccc}
\bottomrule
\multicolumn{1}{c|}{3DMatch / 3DLoMatch} &        & \multicolumn{1}{c}{\textbf{RR} (\%)} & \multicolumn{1}{c}{RE (\degree)} & \multicolumn{1}{c}{TE (cm)} \\ \hline
\multirow{2}{*}{FCGF}                  & w. SOG & 94.45 / 63.73                                         & 2.04 / 4.18                                     & 6.34 / 10.17                \\
                                       & w. FOG & 94.39 / 63.50                                         & 2.04 / 4.05                                     & 6.33 / 10.08                \\ \hline
\multirow{2}{*}{FPFH}                  & w. SOG & 85.89 / 42.28                                         & 2.26 / 4.81                                     & 6.74 / 10.80                \\
                                       & w. FOG & 85.77 / 42.22                                         & 2.25 / 4.59                                     & 6.75 / 10.81                \\ \toprule
\end{tabular}}
\label{tab:graph_order_1}
\end{table}

\begin{table}[h]
\caption{Regstration results on KITTI-10m.}
\centering
\resizebox{0.7\linewidth}{!}{
\begin{tabular}{c|c|ccc}
\bottomrule
KITTI-10m             &        & \multicolumn{1}{c}{\textbf{RR} (\%)} & \multicolumn{1}{c}{RE (\degree)} & \multicolumn{1}{c}{TE (cm)} \\ \hline
\multirow{2}{*}{FCGF} & w. SOG & 98.92                                & 0.32                             & 19.82                       \\
                      & w. FOG & 98.92                                & 0.32                             & 19.83                       \\ \hline
\multirow{2}{*}{FPFH} & w. SOG & 99.10                                & 0.34                             & 7.74                        \\
                      & w. FOG & 99.10                                & 0.34                             & 7.72                        \\ \toprule
\end{tabular}}
\label{tab:graph_order_2}
\end{table}
\begin{table}[h]
\caption{Registration results on KITTI-LC.}
\centering
\resizebox{0.7\linewidth}{!}{
\begin{tabular}{c|c|ccc}
\bottomrule
KITTI-LC                &  & \multicolumn{1}{c}{\textbf{RR} (\%)} & \multicolumn{1}{c}{RE (\degree)} & \multicolumn{1}{c}{TE (cm)} \\ \hline
\multirow{2}{*}{0-10m}  & w. SOG      & 97.05                       &   0.27                     &  7.47                      \\
                        & w. FOG      & 97.05                  & 0.27                   & 7.45                   \\ \hline
\multirow{2}{*}{10-20m} & w. SOG      &  72.46                      &     0.55                   &   15.24                     \\
                        & w. FOG      & 72.37                  & 0.55                   & 15.40                  \\ \hline
\multirow{2}{*}{20-30m} & w. SOG      &  25.95                      &       0.76                 &  22.84                      \\
                        & w. FOG      & 26.43                  & 0.75                   & 21.70                  \\ \toprule
\end{tabular}}
\label{tab:graph_order_3}
\end{table}

\noindent\textbf{Combined with Transformer-based Methods.} We conduct experiments on 3DMatch and 3DLoMatch using the correspondence generated by CoFiNet~\cite{yu2021cofinet} and GeoTrans~\cite{qin2023geotransformer}. We apply the same RE/TE criteria and comparison methods in Sect.~\ref{sec:indoor}. Following~\cite{qin2023geotransformer}, point cloud pairs from adjacent frames are excluded from the evaluation. As shown in Table~\ref{tab:ext-3dmatch}, on the 3DMatch dataset, the quality of correspondences generated by the transformer-based methods is already quite high. Hence, the performance improvements brought by all compared methods are relatively limited. However, HyperGCT has achieved the largest improvements of all. Specifically, HyperGCT boosts CoFiNet by 1.56$\%$, reaching a recall of 93.04$\%$, and improves GeoTrans by 0.54$\%$, achieving a recall of 94.99$\%$. As indicated in Table~\ref{tab:ext-3dlomatch}, on the 3DLoMatch dataset, due to the limited quality of the matches produced by the transformer-based methods, all the compared approaches show significant performance gains. HyperGCT improves CoFiNet by 6.96$\%$, reaching a recall of 70.63$\%$, and increases GeoTrans by 1.50$\%$, achieving a recall of 78.33$\%$.  

\begin{table}[h]
\centering
\caption{Registration results on 3DMatch.}
\resizebox{\linewidth}{!}{
\begin{tabular}{l|ccc|ccc}
\bottomrule
\multirow{2}{*}{} & \multicolumn{3}{c|}{CoFiNet~\cite{yu2021cofinet}} & \multicolumn{3}{c}{GeoTrans~\cite{qin2023geotransformer}} \\ \cline{2-7} 
        & \textbf{RR} ($\%$)          & RE ($\degree$)       & TE ($cm$)       & \textbf{RR} ($\%$)     & RE ($\degree$)       & TE ($cm$)           \\ \hline
Origin &      91.48    &    2.59     &     8.19    & 94.45         &  1.85       &  \underline{6.11}       \\ \hline
SC$^2$-PCR~\cite{chen2023sc}                &    \underline{92.89}      &       2.16  &    6.93     &    94.61      &   \underline{1.84}      &     6.14    \\
MAC~\cite{yang2024mac}               &     {92.73}     & 2.22        &   \textbf{6.55}      &     94.53     &   1.92      &    \textbf{5.74}     \\ \hline
PointDSC~\cite{bai2021pointdsc}          &     92.65     &  \underline{2.15}       &   6.94      &    94.53      &     \underline{1.84}    &    6.14     \\
PG-Net~\cite{wang2023pg}            &     92.65     & \underline{2.15}        &     6.94 &    94.61      &   \underline{1.84}      & 6.15        \\
VBReg~\cite{jiang2023robust}             &    \underline{92.89}      &    2.16    &   6.94      &     \underline{94.68}     &   \underline{1.84}      &   6.15      \\
Hunter~\cite{yao2023hunter}            &     91.63     &     \underline{2.15}    &   6.94      &    \underline{94.68}      &   1.85      &     6.19    \\
3DPCP~\cite{wang20243dpcp}             &     92.18     &    \textbf{2.09}     &    \underline{6.82}     &    94.61      &   \textbf{1.83}      &     6.25    \\
HyperGCT          &     \textbf{93.04}     &    \underline{2.15}     &   6.94      &    \textbf{94.99}     &    \underline{1.84}     &   6.16      \\ \toprule
\end{tabular}}
\label{tab:ext-3dmatch}
\end{table}
\begin{table}[h]
\centering
\caption{Registration results on 3DLoMatch.}
\resizebox{\linewidth}{!}{
\begin{tabular}{l|ccc|ccc}
\bottomrule
\multirow{2}{*}{} & \multicolumn{3}{c|}{CoFiNet~\cite{yu2021cofinet}} & \multicolumn{3}{c}{GeoTrans~\cite{qin2023geotransformer}} \\ \cline{2-7} 
        & \textbf{RR} ($\%$)          & RE ($\degree$)       & TE ($cm$)       & \textbf{RR} ($\%$)     & RE ($\degree$)       & TE ($cm$)           \\ \hline
Origin &     63.67     &   4.20      &     11.47    &  76.83        &    2.84     &   8.69      \\ \hline
SC$^2$-PCR~\cite{chen2023sc}                &  69.00        &    3.40     &    9.73     &     77.58     &   2.86      &    8.78     \\
MAC~\cite{yang2024mac}               &    \underline{70.51}      &    3.54     &    9.82     &    \textbf{78.33}      &    3.01     &   8.82      \\ \hline
PointDSC~\cite{bai2021pointdsc}          &      68.37    &  3.38     &   9.75      &    77.11      &    2.83     &     \underline{8.73}    \\
PG-Net~\cite{wang2023pg}            &     68.95     &  3.39       &   9.74      &     77.52     &    2.86     &    8.80     \\
VBReg~\cite{jiang2023robust}             &   70.34       &     3.40    &   9.79      &     \underline{78.04}     &   2.83      &    8.77     \\
Hunter~\cite{yao2023hunter}            &     65.82     &     \underline{3.36}    &   \underline{9.70}      &   77.35       &   2.87      &     8.90    \\
3DPCP~\cite{wang20243dpcp}             &     67.56     &   \textbf{3.21}      &    \textbf{9.43}     &    76.53      &    \textbf{2.75}     &    \textbf{8.61}     \\
HyperGCT          &     \textbf{70.63}     &     3.39    &   9.84      &    \textbf{78.33}      &     \underline{2.82}    &    8.74     \\ \toprule
\end{tabular}}
\label{tab:ext-3dlomatch}
\end{table}

\noindent\textbf{Model Complexity.} In terms of practical benefits, our method achieves consistent registration, robustness, and generalization advantages. Furthermore, as shown in Tables~\ref{tab:3dmatch} and ~\ref{tab:3dlomatch}, our runtime ($\sim$0.2s) is comparable to VBReg (CVPR 2023), and our model has fewer parameters than both PointDSC (CVPR 2021) and VBReg (Table~\ref{tab:param}). Therefore, the complexity introduced is well justified by the gain in performance and flexibility.
\begin{table}[h]
\caption{Parameter scale of compared learning-based methods.}
\resizebox{\linewidth}{!}{
\begin{tabular}{l|cccccc}
\bottomrule
Method       & PointDSC & PG-Net & VBReg & Hunter & 3DPCP & \textbf{Ours} \\ \hline
\# params (M) & 1.05     & 0.96   & 3.25  & 0.01   & 0.27  & 0.98     \\ \toprule
\end{tabular}}
\label{tab:param}
\end{table}

\noindent\textbf{Inductive Biases.} Our entire architecture is designed to incorporate inductive biases beyond the non-local component, in order to effectively learn high-order geometric constraints. Specifically: 1) We leverage SOG, constructed using dynamic compatibility thresholds, as a strong geometric prior. This helps the model adapt to varying input distributions by injecting structured information directly into the learning process. 2) The backbone of HyperGCT is a graph neural network, which inherently encodes a relational inductive bias. Its shared update functions over vertices and hyperedges allow it to generalize across graphs of different sizes and connectivity patterns. 3) Our multi-layer architecture gradually reduces the ratio of node–hyperedge associations across layers. This enables broad information propagation in shallow layers, while enforcing more reliable, selective message passing in deeper layers. We believe these combined inductive biases contribute significantly to the robustness and generalization capability of our method.

\noindent\textbf{Generalization.} Our method's improved generalization comes from two aspects. First, unlike conventional correspondence-based approaches that depend heavily on dataset-specific discriminative features, our method treats relationships across different data modalities as a generalized, dataset-agnostic hypergraph structure. This helps the front-end network mitigate biases tied to specific data modalities, enhancing generalization and robustness. Second, unlike traditional outlier rejection strategies that rely solely on learned discriminative features, our back-end solver leverages learned geometric constraints to explore the solution space in a more nuanced manner systematically. This approach avoids the limitations of feature-centric outlier filtering and instead captures a broader range of potential solutions, significantly enhancing hypothesis generation and verification processes. 

\section{Dataset Statistics}\label{sec:dataset}
We present the details for all datasets in Table~\ref{tab:dataset}, including the average number and inlier ratios for FPFH matches.
\begin{table}[h]
    \centering
    \caption{Dataset information.}
    \resizebox{\linewidth}{!}{
    \begin{tabular}{l|c|c|c|c|c}
\bottomrule
Dataset                                        & Type                     & Modality               & \# Pairs & \# Matches            & Inlier ratio \\ \hline
3DMatch~\cite{zeng20173dmatch}                 & Indoor                   & RGB-D                  & 1623     & 4710                  & 6.84\%       \\ \hline
3DLoMatch~\cite{huang2021predator}             & Indoor                   & RGB-D                  & 1781     & 4653                  & 1.68\%       \\ \hline
KITTI-10m~\cite{qiao2024g3reg}                 & Outdoor                  & LiDAR                  & 554      & 8000                  & 4.52\%       \\ \hline
\multirow{3}{*}{KITTI-LC~\cite{qiao2024g3reg}} & \multirow{3}{*}{Outdoor} & \multirow{3}{*}{LiDAR} & 914      & \multirow{3}{*}{8000} & 6.25\%       \\
                                               &                          &                        & 1151     &                       & 1.94\%       \\
                                               &                          &                        & 1260     &                       & 0.81\%       \\ \hline
ETH~\cite{pomerleau2012challenging}            & Outdoor                  & LiDAR                  & 713      & 5000                  & 0.84\%       \\ \toprule
\end{tabular}}
    \label{tab:dataset}
\end{table}

\section{Visualizations}
\begin{figure*}[t]
\centering
    \includegraphics[width=\linewidth]{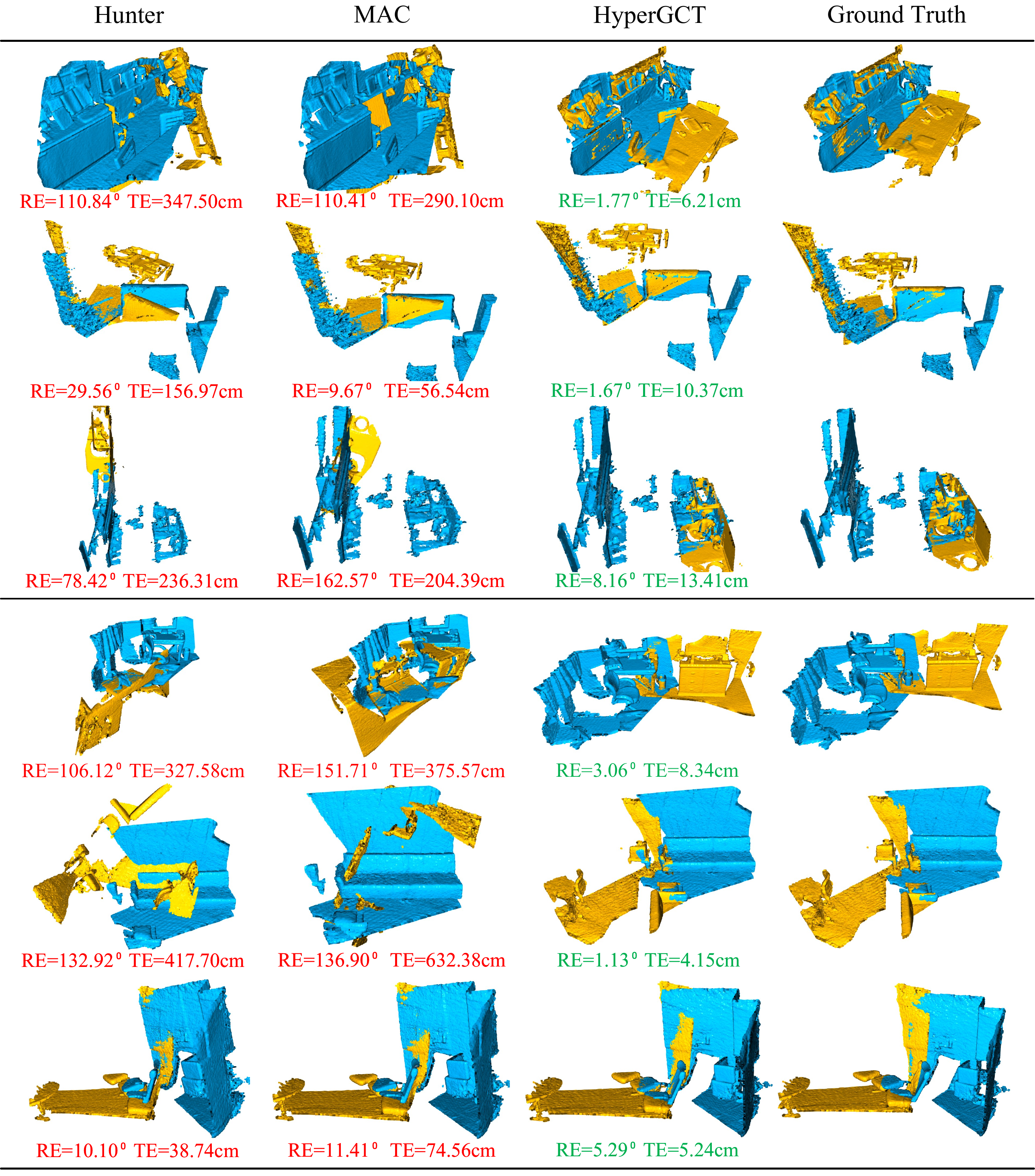}
    \caption{Visualizations of registration results. Rows 1-3 are from 3DMatch, Rows 4-6 are from 3DLoMatch.}
    \label{fig:supp2}
\end{figure*}

\begin{figure*}[t]
\centering
    \includegraphics[width=0.9\linewidth]{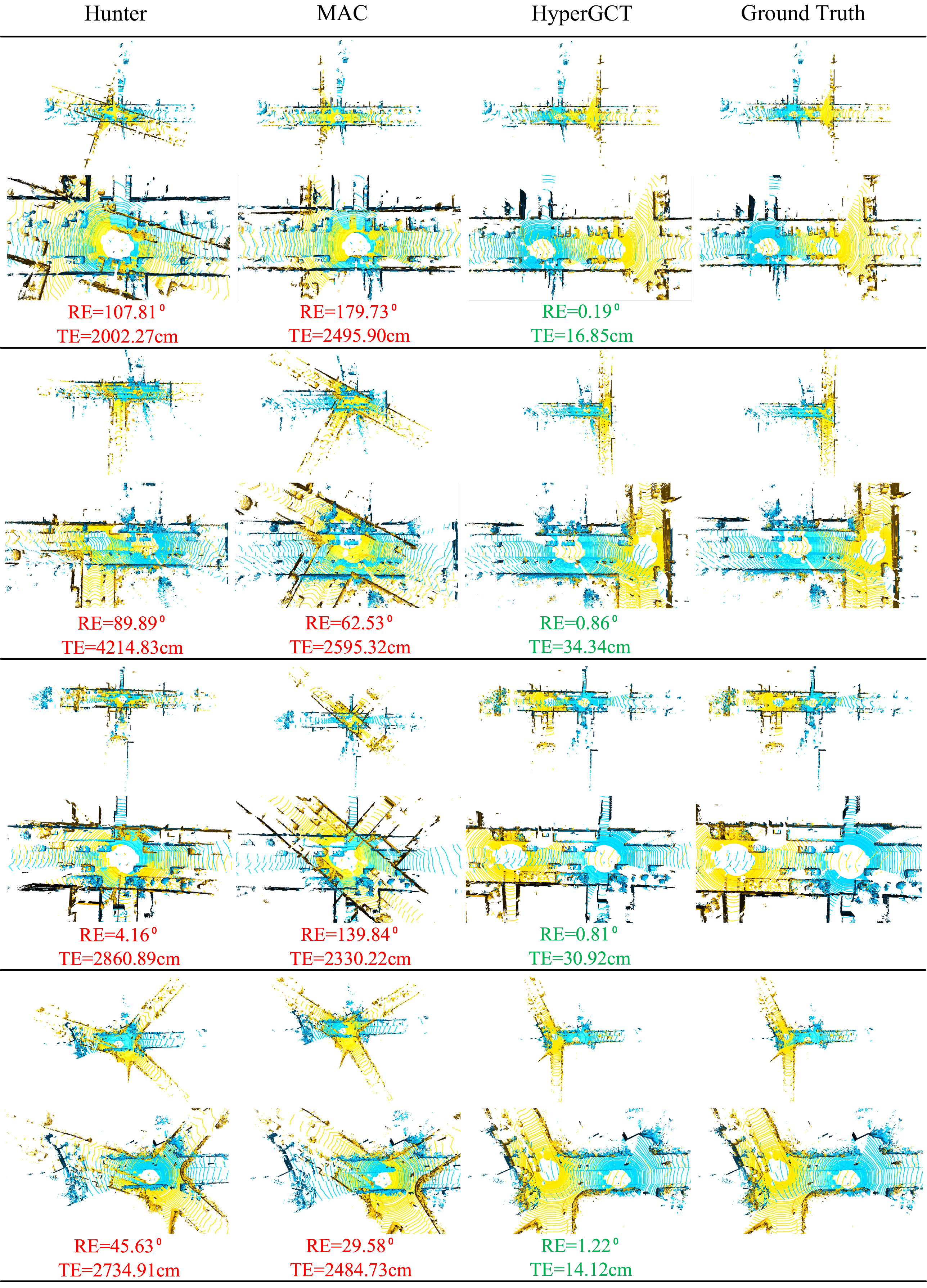}
    \caption{Visualizations of registration results on KITTI-LC.  Each visualization provides both a global view (upper row) and a detailed local perspective (lower row).}
    \label{fig:supp3}
\end{figure*}

\begin{figure*}[t]
\centering
    \includegraphics[width=\linewidth]{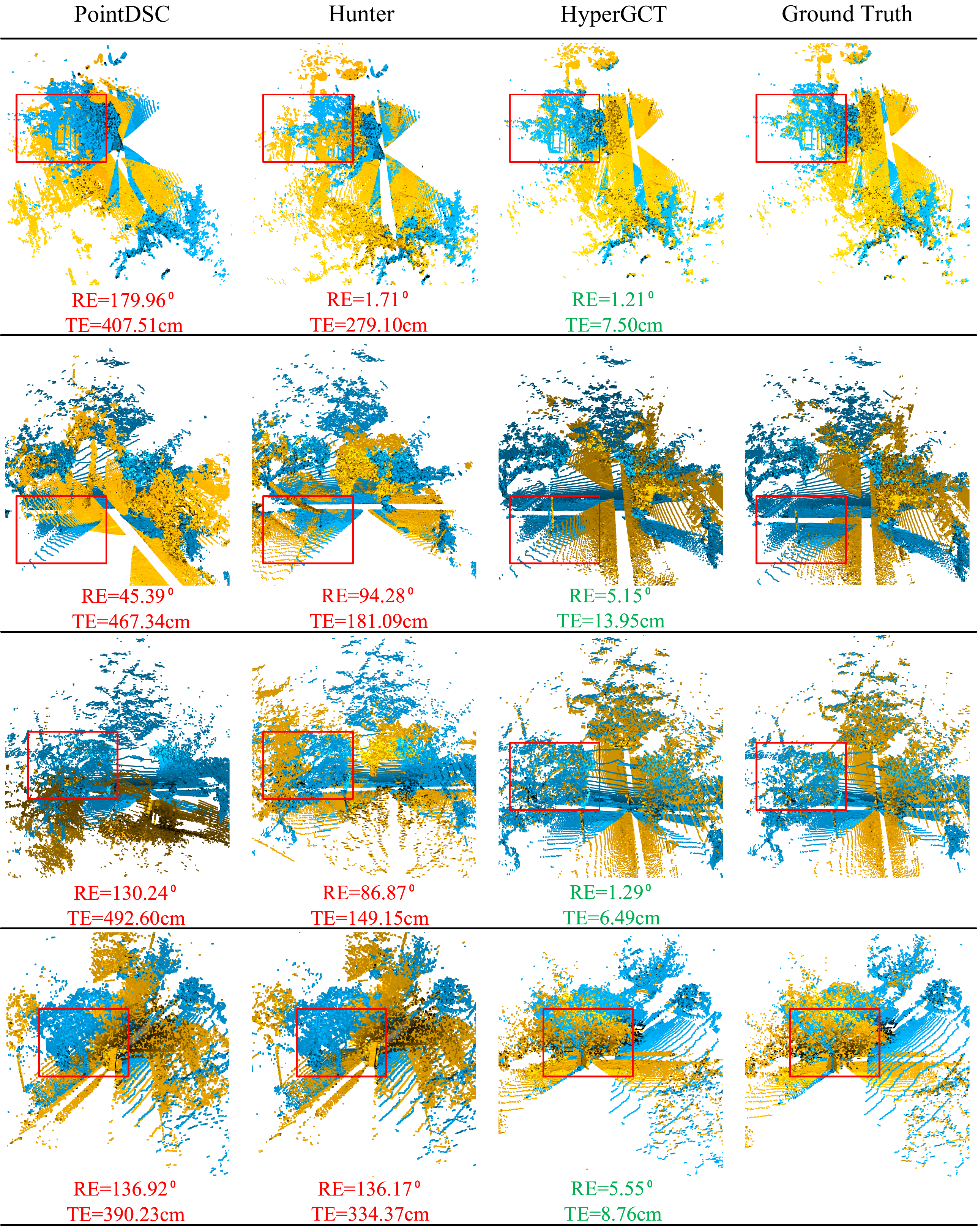}
    \caption{Visualizations of generalization results on ETH. For each set of results, we highlight the same point cloud region with a \textcolor{red}{red box} to compare differences across methods.}
    \label{fig:supp4}
\end{figure*}

We provide visualizations of registration results on the indoor scene 3DMatch (Fig.~\ref{fig:supp2}), outdoor scene KITTI (Fig.~\ref{fig:supp3}), and outdoor scene ETH (Fig.~\ref{fig:supp4}). Our approach accurately registers the data where PointDSC, Hunter, and MAC fail. This demonstrates that HyperGCT offers stronger constraint capability for generating correct transformation hypotheses.

